\pdfoutput=1
\documentclass[11pt]{article}
\usepackage[preprint]{acl}
\usepackage{times}
\usepackage{latexsym}
\usepackage[T1]{fontenc}
\usepackage[utf8]{inputenc}
\usepackage{microtype}
\usepackage{inconsolata}
\usepackage{graphicx}
\usepackage{longtable}
\usepackage{booktabs}
\usepackage{multirow}
\usepackage{colortbl}
\usepackage{tikz}
\usepackage{amsmath}
\usepackage{subcaption}
\usepackage{listings}
\usepackage{float}
\usepackage{hyperref}

\title{Towards Visuospatial Cognition via Hierarchical Fusion of Visual Experts\thanks{This is a technical report and a draft version. Work in progress.}}

\author{
Qi Feng${}^{1}$\\
${}^{1}$Kyoto University\\
\texttt{feng.qi.45n@st.kyoto-u.ac.jp}\\
\url{https://github.com/nkkbr/ViCA}
}

\begin{document}
\maketitle
\begin{abstract}
While Multimodal Large Language Models (MLLMs) excel at general vision-language tasks, \textit{visuospatial cognition}—reasoning about spatial layouts, relations, and dynamics—remains a significant challenge. Existing models often lack the necessary architectural components and specialized training data for fine-grained spatial understanding. We introduce \textbf{ViCA2} (\textbf{Vi}suospatial \textbf{C}ognitive \textbf{A}ssistant 2), a novel MLLM designed to enhance \textit{spatial reasoning}. \textbf{ViCA2} features a dual vision encoder architecture integrating \texttt{SigLIP} for semantics and \texttt{Hiera} for spatial structure, coupled with a token ratio control mechanism for efficiency. We also developed \textbf{ViCA-322K}, a new large-scale dataset with over 322,000 spatially grounded question-answer pairs for targeted instruction tuning. On the challenging \textbf{VSI-Bench} benchmark, our \textbf{ViCA2-7B} model achieves a state-of-the-art average score of 56.8, significantly surpassing larger open-source models (e.g., \textbf{LLaVA-NeXT-Video-72B}, 40.9) and leading proprietary models (\textbf{Gemini-1.5 Pro}, 45.4). This demonstrates the effectiveness of our approach in achieving strong visuospatial intelligence with a compact model. We release \textbf{ViCA2}, its codebase, and the \textbf{ViCA-322K} dataset to facilitate further research.
\end{abstract}

\section{Introduction}

Recent advances in multimodal large language models (MLLMs) have led to remarkable progress in visual-language tasks such as image captioning, visual question answering (VQA), and video understanding. By aligning vision and language modalities through large-scale pretraining, models such as Flamingo\citep{alayrac2022flamingovisuallanguagemodel}, LLaVA\citep{liu2024visual}, and their successors have shown strong generalization capabilities across a wide range of benchmarks. More recently, models like LLaVA-OneVision\citep{li2024llavaonevision} and LLaVA-NeXT-Video\citep{zhang2024llavanextvideo} have extended these capabilities to multi-image and video scenarios, enabling unified processing of diverse visual inputs via high-resolution token streams, instruction tuning, and stronger visual backbones such as SigLIP\citep{zhai2023sigmoid}.

\begin{figure}[t!]
  \includegraphics[width=\columnwidth]{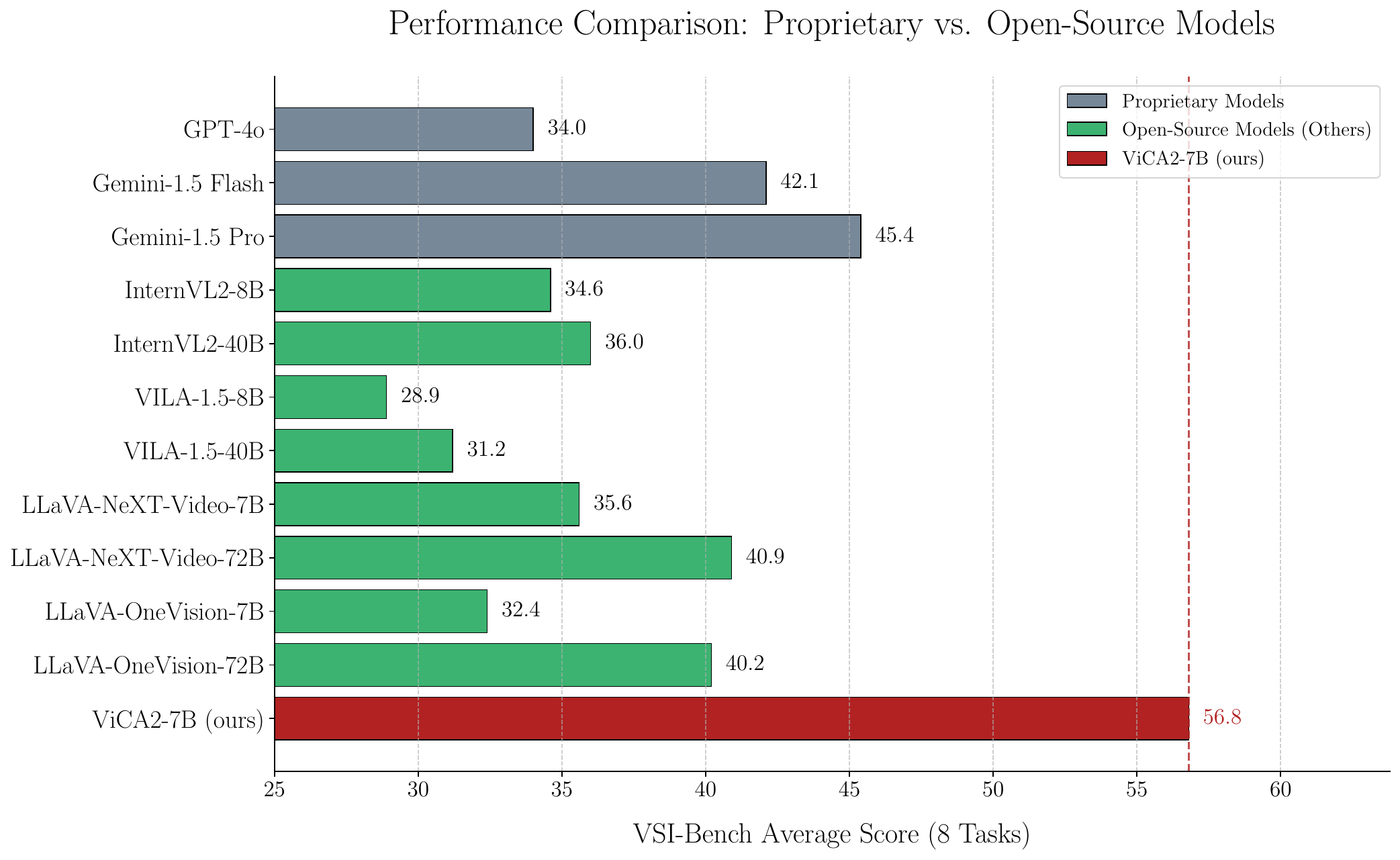}
  \caption{
\textbf{Average performance comparison on VSI-Bench.}
We compare ViCA2-7B with both proprietary API-based models and leading open-source multimodal models on the VSI-Bench benchmark, which evaluates visuospatial reasoning across 8 tasks.
Despite having only 7B parameters, ViCA2-7B significantly outperforms all other models, achieving an average score of \textbf{56.8}, surpassing the best-performing proprietary model (Gemini-1.5 Pro\citep{team2024gemini}, 45.4) and the strongest open-source baseline (LLaVA-NeXT-Video-72B, 40.9) by a large margin.
This highlights the effectiveness of our dual vision encoder design and hierarchical fusion approach for spatial understanding.
}
\label{fig:first_img}
\end{figure}

While these advances are significant, an important aspect of human-level visual intelligence remains largely underexplored: \textit{visuospatial cognition}—the capacity to construct, retain, and reason over spatially coherent internal representations of complex environments. In contrast to general scene understanding, visuospatial reasoning requires fine-grained comprehension of object layouts, spatial relations, temporal order, and geometric attributes. Such capabilities are critical for downstream applications including indoor navigation, assistive robotics, video summarization, and embodied question answering.

Recent work has begun to probe this capacity. \citet{yang2024thinking} introduced VSI-Bench, a challenging benchmark targeting spatial intelligence through a suite of question types over real indoor videos. Their analysis revealed that even state-of-the-art MLLMs like LLaVA-OneVision-72B and LLaVA-NeXT-Video-72B struggle significantly on spatially grounded tasks. Although proprietary models such as Gemini-1.5 Pro show stronger spatial abilities, these systems remain closed-source and resource-heavy. A key bottleneck identified in open models is their reliance on a single semantic vision encoder, which often fails to capture high-resolution spatial layouts due to low input resolutions and an architecture optimized for global semantic alignment. Moreover, the lack of task-specific training data targeting spatial reasoning further limits their capacity to internalize structured spatial knowledge.

\paragraph{Our Contributions}
In this paper, we propose \textbf{ViCA2} (\textbf{Vi}suospatial \textbf{C}ognitive \textbf{A}ssistant 2), a lightweight yet powerful multimodal framework designed to advance visuospatial understanding in both image and video settings. Our approach builds upon \textbf{LLaVA-OneVision} and introduces several key innovations:

\begin{enumerate}
    \item \textbf{Dual Vision Encoder Architecture for Enhanced Spatial Awareness:} We propose a novel dual vision encoder architecture that integrates \texttt{SigLIP} for global semantic abstraction with \texttt{Hiera}—a hierarchical visual encoder from SAM2\citep{ravi2024sam}—for modeling fine-grained spatial structure. The two feature streams are individually projected into the language model's embedding space and fused prior to decoding, enabling joint reasoning over semantics and spatial cues.

    \item \textbf{Efficient Token Ratio Control Mechanism:} We develop a token ratio control mechanism that adjusts frame sampling density, feature extraction depth, and spatial pooling stride for each encoder. This offers explicit control over the balance between semantic abstraction and spatial detail, particularly crucial under memory constraints for handling high-resolution or long-duration visual inputs.

    \item \textbf{State-of-the-Art Performance with a Compact Model:} Our experimental results demonstrate that \textbf{ViCA2-7B}, despite using a relatively compact 7B language model, achieves state-of-the-art performance on the challenging VSI-Bench benchmark (Figure~\ref{fig:first_img}). Specifically, \textbf{ViCA2-7B} significantly outperforms much larger open-source models, including LLaVA-OneVision-72B and LLaVA-NeXT-Video-72B, and even surpasses the proprietary Gemini-1.5~Pro in several spatial reasoning categories, highlighting the efficacy of our architectural and data-centric design choices. 
   
\end{enumerate}

To facilitate further research in visuospatial cognition, we open-source our model weights, codebase, and the \textbf{ViCA-322K} dataset.

\section{Related work}

\subsection{Multimodal Large Language Models}

Multimodal large language models (MLLMs) integrate vision and language understanding within a unified architecture by coupling pretrained language models with visual encoders through lightweight projection modules. Early models such as Flamingo and MiniGPT-4~\citep{zhu2023minigpt} demonstrated the effectiveness of aligning image features with language inputs. Recent open-source models like LLaVA-OneVision and Video-LLaMA~\citep{zhang2023video} extend this capability to multi-image and video understanding using variable-length visual sequences. Meanwhile, proprietary systems like Gemini 2.5 Pro~\citep{kavukcuoglu2025gemini25} and GPT-4V~\citep{openai2023gpt4v} exhibit strong performance on vision-language tasks, albeit with closed architectures. 

\subsection{Visual Encoders for Multimodal Learning}

\begin{figure*}[t]
  \centering
  \includegraphics[width=\textwidth]{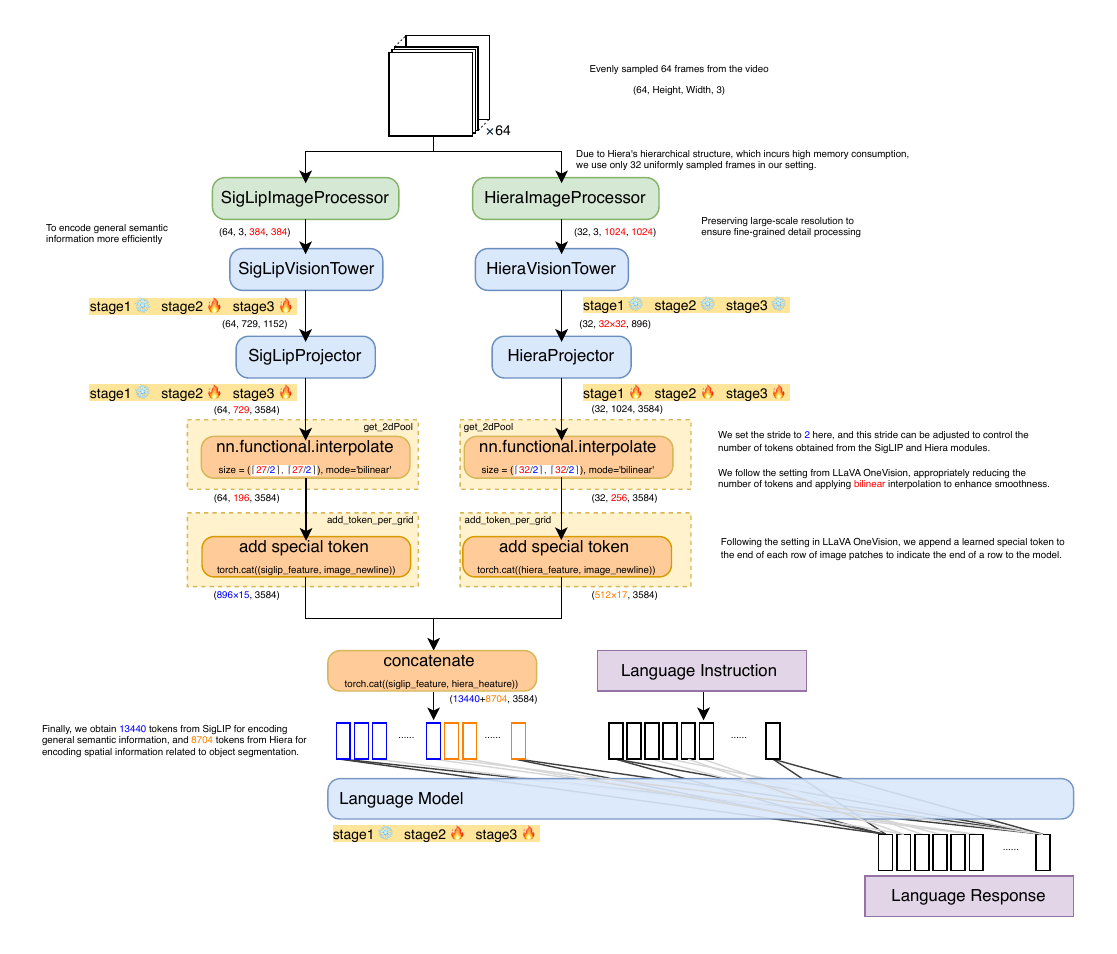} 
  \caption{\textbf{An overview of the ViCA2 architecture.} The model integrates SigLIP for global semantic understanding and Hiera for spatial structure encoding, with a token ratio control mechanism to balance expressiveness and memory efficiency. A complete version of the architecture diagram is provided in the Appendix(Figure~\ref{fig:vica2_arch_complete}).}
  \label{fig:vica2_arch}
\end{figure*}

Most MLLMs rely on visual encoders pretrained on image-text contrastive learning, such as CLIP~\citep{radford2021learning}, EVA-CLIP~\citep{sun2023eva}, or SigLIP. While effective for global semantics, these models are typically trained on low-resolution inputs, limiting their ability to capture fine-grained spatial details. Recent solutions adopt higher resolutions or tiling strategies (e.g., LLaVA-UHD~\citep{guo2024llava}, InternLM-XComposer~\citep{zhang2023internlm}) to improve spatial coverage, but these approaches do not fundamentally enhance spatial reasoning. In contrast, encoders pretrained on dense prediction tasks—such as Hiera~\citep{ryali2023hiera}, a hierarchical vision backbone used in SAM2—offer stronger spatial sensitivity. Our ViCA2 model leverages a dual-encoder design combining SigLIP for global semantics and Hiera for spatial structure, with independent projection heads and token ratio control to balance expressiveness and efficiency.

\subsection{Visuospatial Reasoning in MLLMs}

While MLLMs excel at general vision-language tasks, they remain limited in visuospatial reasoning, which involves understanding object positions, spatial relations, and temporal dynamics. VSI-Bench highlights this gap by evaluating models on spatial questions grounded in real-world indoor videos. Even advanced models such as LLaVA-OneVision-72B and Gemini-1.5 Pro struggle with consistent spatial understanding due to inadequate training supervision and architectural constraints.

\section{ViCA2}

\subsection{The Overview}

LLaVA-OneVision is a state-of-the-art open-source large multimodal model that unifies single-image, multi-image, and video understanding within a single architecture. It integrates a strong visual encoder, \texttt{SigLIP}, with a powerful language model, \texttt{Qwen-2}, and employs a lightweight MLP-based projection layer to bridge vision and language modalities. The model achieves high performance across a wide range of visual-language benchmarks by leveraging a unified representation strategy and a large-scale, high-quality training corpus. Its success demonstrates the effectiveness of multi-stage instruction tuning and the benefits of treating visual input—regardless of type—as sequences of visual tokens.

Building on LLaVA-OneVision, our proposed model, ViCA2(Figure~\ref{fig:vica2_arch}), extends the multimodal capabilities by introducing a second vision encoder derived from \texttt{Hiera}, the hierarchical visual backbone used in SAM2. While LLaVA-OneVision utilizes SigLIP to capture global semantic features, our ViCA2 further incorporates the spatially structured features extracted by \texttt{Hiera} to enhance fine-grained understanding of object-centric and region-level information. The outputs from both vision encoders are individually projected into the language model's embedding space via two separate linear projectors, and then fused before being consumed by the decoder. This dual-encoder architecture enables the model to jointly reason over high-level semantics and spatial visual cues, significantly improving its performance on spatially grounded tasks.

\subsection{Dual Vision Encoders for Spatially-Grounded Understanding}

We employ \texttt{SigLIP} as one of the vision encoders in our model to extract high-level semantic representations from visual inputs. SigLIP benefits from contrastive pretraining on large-scale image-text pairs. It produces compact and expressive visual embeddings that align well with natural language, making it highly effective for tasks such as image captioning and general visual question answering. However, models employing a CLIP-like contrastive learning paradigm, including \texttt{SigLIP}, are primarily optimized for global semantic alignment rather than fine-grained spatial reasoning. This limitation is particularly evident in phenomena such as "CLIP-blind pairs" \citep{tong2024eyes}, where visually distinct images are mapped to similar embeddings due to semantic proximity. Furthermore, the relatively low-resolution input constraints (e.g., 336×336 or 512×512) make it difficult for these models to preserve detailed local features, which are essential for dense OCR, visual grounding, and region-level understanding.

To address these limitations, we introduce a second vision encoder based on \texttt{Hiera}, the hierarchical visual backbone from SAM2.1 (specifically, the \texttt{sam2.1\_hiera\_b+} variant), to complement the \texttt{SigLIP} encoder. Unlike the ViT-H encoder used in the original SAM, \texttt{Hiera} adopts a multi-stage hierarchical structure that extracts multiscale visual features through progressive downsampling, similar in spirit to classical feature pyramids. Pretrained with Masked Autoencoding (MAE), \texttt{Hiera} is particularly well-suited for spatially dense tasks, enabling robust recognition of objects and regions that vary in size and location. This is especially beneficial for video or multi-image scenarios, where spatial consistency and resolution variance are critical. Compared to ViT-H, \texttt{Hiera} not only improves segmentation and localization accuracy but also reduces computation time significantly, making it a more efficient and spatially aware encoder for large-scale multimodal models.

In our architecture, the outputs from both \texttt{SigLIP} and \texttt{Hiera} are processed through separate linear projectors to align their feature spaces with the language model's embedding space. The projected features are then concatenated along the sequence dimension before being fed into the language decoder. This dual-encoder setup allows our model to jointly reason over global semantic content and spatially grounded visual cues, thereby enhancing its spatial understanding capabilities. By leveraging \texttt{SigLIP}'s strength in semantic abstraction and \texttt{Hiera}'s ability to model fine-grained spatial structures, the model is better equipped to tackle challenging tasks that require object-level reasoning, region grounding, and multi-object spatial relation understanding across both images and videos.

\subsection{Video Feature Extraction and Dual Encoder Token Balancing}

To effectively process video inputs, we uniformly sample $N_{\text{total}} = 64$ frames from each video. These frames are fed into two distinct visual encoders: \texttt{SigLIP} and \texttt{Hiera}. While \texttt{SigLIP} processes all 64 frames to extract high-level semantic features, the computational cost of \texttt{Hiera}—particularly its hierarchical structure and high spatial resolution in early stages—necessitates a more selective approach. To strike a balance between expressiveness and memory usage, we feed only $N_{\text{hiera}} = 32$ uniformly sampled frames into \texttt{Hiera}.

Each frame destined for \texttt{Hiera} is first preprocessed using the SAM2.1 \texttt{HieraImageProcessor} and then passed through the \texttt{Hiera} backbone, which comprises four hierarchical stages. As the stage depth increases, spatial resolution is progressively reduced while channel capacity increases. Early stages preserve fine-grained visual details such as edges and textures, whereas later stages encode global semantic structures such as object identities and scene layouts. In our configuration, we extract features from Stage 4, the deepest layer, which captures the most abstract and semantically rich information. These features are then passed through a 2D spatial pooling module with stride $S_{\text{pool}} = 2$, implemented using bilinear interpolation, following the token downsampling strategy used in LLaVA-OneVision. This reduces the token count per frame while preserving the spatial continuity of features, enabling a higher number of frames to be processed within the same memory budget.

To further enhance spatial layout awareness, we append a learned row-wise special token at the end of each row in the resulting 2D token grid, providing the model with positional cues about horizontal boundaries. After this transformation, we obtain $T_{\text{siglip}} = 13,440$ tokens from SigLIP and $T_{\text{hiera}} = 8,704$ tokens from Hiera. These two token sequences are concatenated along the sequence dimension and fed into the language decoder. The token ratio $T_{\text{siglip}} : T_{\text{hiera}} = 1.54$ reflects a deliberate design choice to balance global semantics and spatial precision.

We introduce a token ratio control strategy based on a configurable triplet $(N_{\text{hiera}}, S_{\text{stage}}, S_{\text{pool}})$, which respectively denote: the number of frames passed to \texttt{Hiera}, the selected stage for feature extraction, and the spatial pooling stride. This mechanism enables fine-grained control over the balance between semantic and spatial representations. A high proportion of \texttt{SigLIP} tokens tends to bias the model toward global semantic alignment, while an excess of \texttt{Hiera} tokens increases spatial sensitivity but may dilute language grounding. Our chosen configuration $(32, 4, 2)$ offers a favorable trade-off, enhancing the model's ability to perform spatially grounded reasoning tasks without exceeding hardware limitations.

\section{Training}

\subsection{Initialization of Model Parameters}

\begin{table*}[t]
  \centering
  \resizebox{\textwidth}{!}{%
  \begin{tabular}{l|cccc}
  \toprule
  \textbf{} & \textbf{Stage 1} & \textbf{Stage 2} & \textbf{Stage 3} & \textbf{Thinking} \\
  \midrule
  \textbf{Dataset} & LLaVA-CC3M-Pretrain-595K & LLaVA-OneVision-Data & ViCA-322K & ViCA-Thinking-2.68K \\
  \textbf{Training Data Samples} & 595{,}375 & 279{,}353 & 322{,}003 & 2{,}680 \\
  \textbf{Trainable Module} & Hiera Projection Layer & \multicolumn{3}{c}{Full Model (excluding Hiera Module)} \\
  \textbf{Trainable Parameters} & 16M & 8.04B & 8.04B & 8.04B \\
  \textbf{Learning Rate} & 1e-3 & 1e-5 & 1e-5 & 1e-5 \\
  \textbf{Epochs} & 1 & 1 & 1 & 1 \\
  \textbf{DeepSpeed Stage} & ZeRO-2 & ZeRO-3 & ZeRO-3 & ZeRO-3 \\
  \bottomrule
  \end{tabular}
  }
  \caption{\textbf{Training configuration across all four stages of our hierarchical vision-language model.} Stage 1 pretrains only the Hiera projection layer, while stages 2-4 fine-tune the full model excluding the Hiera module.}
  \label{tab:training_stages}
  \end{table*}

Our model is initialized based on the publicly released checkpoint \texttt{lmms-lab/LLaVA-Video-7B-Qwen2} on Hugging Face. Specifically, the language model is initialized from \texttt{Qwen2-7B}, and the primary vision encoder, SigLIP, is initialized from the pretrained weights of \texttt{google/siglip-so400m-patch14-384}. To enhance spatial reasoning capabilities, we incorporate a second vision encoder, Hiera, which is initialized from the \texttt{Hiera} module in \texttt{facebook/sam2.1-hiera-base-plus}. The associated linear projector that maps \texttt{Hiera} features into the language model's embedding space is randomly initialized using Xavier initialization. The projected features from both \texttt{SigLIP} and \texttt{Hiera} are directly concatenated along the sequence dimension without introducing any additional parameterized fusion module.

\subsection{Training Strategy and Data}

We adopt a three-stage training strategy to progressively adapt our model—particularly the newly introduced \texttt{Hiera} encoder and its associated projection module—for spatially grounded multimodal understanding. Each stage targets specific alignment challenges and uses carefully selected datasets with different scales and characteristics.

In the first stage, we focus on calibrating the newly added \texttt{Hiera} projector, a two-layer MLP that is randomly initialized and initially incompatible with the language model's input space. To facilitate coarse visual-language alignment, we employ the \texttt{liuhaotian/LLaVA-CC3M-Pretrain-595K} dataset, which contains large-scale image-text pairs with relatively simple and descriptive captions. This dataset is well-suited for warming up the \texttt{Hiera} pathway without introducing overly complex semantics. During this stage, we freeze all other model components and only update the \texttt{Hiera} projector using a learning rate of 1e-3. This step allows the model to establish a basic alignment between \texttt{Hiera}-derived features and the language model input, laying the foundation for downstream integration.

The second stage aims to restore the full multimodal capability of the base model, \texttt{LLaVA-Video-7B-Qwen2}, which has already been fine-tuned on the full \texttt{lmms-lab/LLaVA-OneVision-Data} corpus. Since the introduction of \texttt{Hiera} and its randomly initialized projector may disrupt previously learned representations, we propose to refine this pathway using only a 10\% subset of the original dataset. In our implementation, this corresponds to 279,353 training samples. This lightweight stage serves to re-align the newly added features from \texttt{Hiera} with those of \texttt{SigLIP} and the language model, without the need for full-scale retraining. The learning rate is set to 1e-5, and during this stage we update \texttt{SigLIP}, both vision projectors, and the language model, while keeping the \texttt{Hiera} encoder frozen due to memory constraints.

In the third stage, we conduct targeted fine-tuning on the ViCA-322K dataset, which we curated and publicly released on Hugging Face. This dataset comprises over 300K spatially grounded question-answer pairs derived from indoor videos, covering a wide spectrum of spatial reasoning tasks, ranging from direct spatial knowledge questions—such as room size estimation, relative distances between objects, and temporal ordering of object appearances—to more complex spatial inference and relational understanding. This stage is designed to enhance the model's spatial perception and reasoning abilities, which are critical for indoor visual-language applications. The training configuration remains the same as in the second stage, with a learning rate of 1e-5, updating all major trainable components except the frozen \texttt{Hiera} backbone.

To manage computational costs, we use DeepSpeed ZeRO-2 with offloading in the first stage, which involves relatively few trainable parameters. In the second and third stages, we adopt DeepSpeed ZeRO-3 with offloading to support full fine-tuning under our hardware constraints. An overview of the training schedule, data composition, and optimization setup is summarized in Table~\ref{tab:training_stages}.

\subsection{Stage-wise Results and Observations}
\label{ssec:stage-wise_results_and_observations}

\begin{table*}[t]
    \small
    \centering
    \resizebox{\textwidth}{!}{ 
    \begin{tabular}{l|c|cccc|cccc}
    \toprule
    \textbf{Method} & \textbf{Average} & \multicolumn{4}{c|}{\textbf{Numerical Answer}} & \multicolumn{4}{c}{\textbf{Multiple-Choice Answer}} \\
      &  & Obj. Count & Abs. Dist. & Obj. Size & Room Size & Rel. Dist. & Rel. Dir. & Route Plan & Appr. Order \\
    \midrule
    \rowcolor{gray!20} \multicolumn{10}{l}{\textit{Proprietary Models (API)}} \\
    GPT-4o & 34.0 & 46.2 & 5.3 & 43.8 & 38.2 & 37.0 & 41.3 & 31.5 & 28.5 \\
    Gemini-1.5 Flash & 42.1 & 49.8 & 30.8 & 53.5 & 54.4 & 37.7 & 41.0 & 31.5 & 37.8 \\
    Gemini-1.5 Pro & 45.4 & 56.2 & 30.9 & 64.1 & 43.6 & 51.3 & \cellcolor{gray!40}46.3 & 36.0 & 34.6 \\
    \midrule
    \rowcolor{gray!20} \multicolumn{10}{l}{\textit{Open-source Models}} \\
    InternVL2-8B & 34.6 & 23.1 & 28.7 & 48.2 & 39.8 & 36.7 & 30.7 & 29.9 & 39.6 \\
    InternVL2-40B & 36.0 & 34.9 & 26.9 & 46.5 & 31.8 & 42.1 & 32.2 & 34.0 & 39.6 \\
    VILA-1.5-8B & 28.9 & 17.4 & 21.8 & 50.3 & 18.8 & 32.1 & 34.8 & 31.0 & 24.8 \\
    VILA-1.5-40B & 31.2 & 22.4 & 24.8 & 48.7 & 22.7 & 40.5 & 25.7 & 31.5 & 32.9 \\
    LLaVA-NeXT-Video-7B & 35.6 & 48.5 & 14.0 & 47.8 & 24.2 & 43.5 & \textbf{42.4} & 34.0 & 30.6 \\
    LLaVA-NeXT-Video-72B & 40.9 & 48.9 & 22.8 & 57.4 & 35.3 & 42.4 & 36.7 & 35.0 & 48.6 \\
    LLaVA-OneVision-7B & 32.4 & 47.7 & 20.2 & 47.4 & 12.3 & 42.5 & 35.2 & 29.4 & 24.4 \\
    LLaVA-OneVision-72B & 40.2 & 43.5 & 23.9 & 57.6 & 37.5 & 42.5 & 39.9 & 32.5 & 44.6 \\
    ViCA2-7B (\textbf{ours}) & \cellcolor{gray!40}\textbf{56.8(+11.4)} & \cellcolor{gray!40}\textbf{65.7(+9.5)} & \cellcolor{gray!40}\textbf{51.0(+20.1)} & \cellcolor{gray!40}\textbf{75.5(+11.4)} & \cellcolor{gray!40}\textbf{71.4(+17.0)} & \cellcolor{gray!40}\textbf{51.6(+0.3)} & 34.6 & \cellcolor{gray!40}\textbf{38.1(+2.1)} & \cellcolor{gray!40}\textbf{66.5(+17.9)} \\
    \bottomrule
    \end{tabular}
    }
    \caption{
    \textbf{Comparison of different models on VSI-Bench.} 
    Our ViCA2-7B achieves the best performance across most metrics. 
    \colorbox{gray!20}{Gray shading} indicates the best overall performance among all models, including 72B-scale and proprietary models, 
    while \textbf{bold font} indicates the best performance among open-source models with 7B/8B scale. 
    The numbers in parentheses (e.g., +20.1) represent the improvement margins over the best-performing model excluding ViCA2-7B. 
    }
    
    \label{tab:comparison}
    \end{table*}

To better understand the effectiveness of our three-stage training strategy, we analyze both the \textit{training dynamics} and the \textit{output performance} of the model across four checkpoints: (1) the randomly initialized projector checkpoint, and (2--4) the model after each of the three fine-tuning stages.

We first examine the training loss curves. In both the first and third stages, the loss decreases sharply at the beginning of training and stabilizes thereafter---indicating successful learning and convergence. In contrast, the second stage exhibits only a mild reduction in training loss. We attribute this to the fact that the base model, \texttt{LLaVA-Video-7B-Qwen2}, had already been thoroughly fine-tuned on the same data used in this stage, leaving limited room for further optimization. These trends are visualized in Appendix Figure~\ref{fig:qualitative_4x3}.

\begin{figure}[t]
  \centering
  \includegraphics[width=0.95\linewidth]{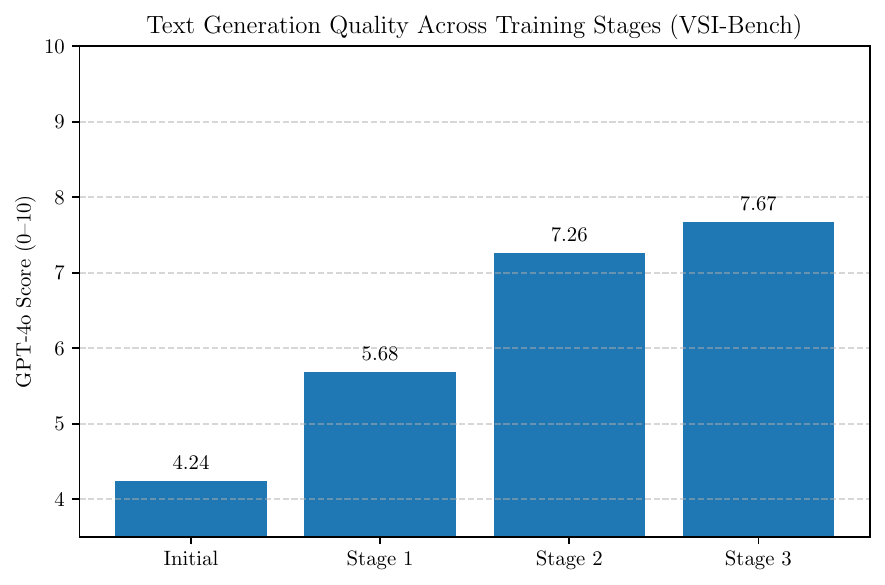}
  \caption{\textbf{Average gpt-4.1-mini scores (0--10) for in-detail descriptions generated by each of the four checkpoints on VSI-Bench.}}
  \label{fig:vsi-eval}
\end{figure}

To assess the impact of each stage on text generation quality, we conduct a controlled evaluation using the \textbf{VSI-Bench} dataset, which consists of 288 indoor video clips. For each video, we prompt the model to generate a \textit{detailed description}. We compare the outputs from the four checkpoints using \textbf{gpt-4.1-mini}\citep{openai2025gpt41} as a judge. Gpt-4.1-mini assigns a quality score in the range $[0, 10]$ for each output based on informativeness, fluency, and spatial relevance. Since some undertrained models may fail to generate the end-of-sequence (EOS) token properly, we set a maximum generation length of 4096 tokens to ensure termination.

As shown in Figure~\ref{fig:vsi-eval}, we observe a \textit{progressive and significant improvement} in average generation score across the four checkpoints. Notably, the model fine-tuned with the \textbf{ViCA-322K} dataset in the final stage achieves the highest score, confirming that instruction tuning on a spatially grounded, mixed-format dataset not only enhances \textit{spatial reasoning}, but also boosts the overall \textit{language generation capability} of the model.\footnote{See Appendix Figure~\ref{fig:qualitative_comparisons_of_the_four_checkpoints} for qualitative comparisons of the four checkpoints on the same video sample.}

\section{Experiements Results}

\begin{figure*}[t]
\centering
\includegraphics[width=\textwidth]{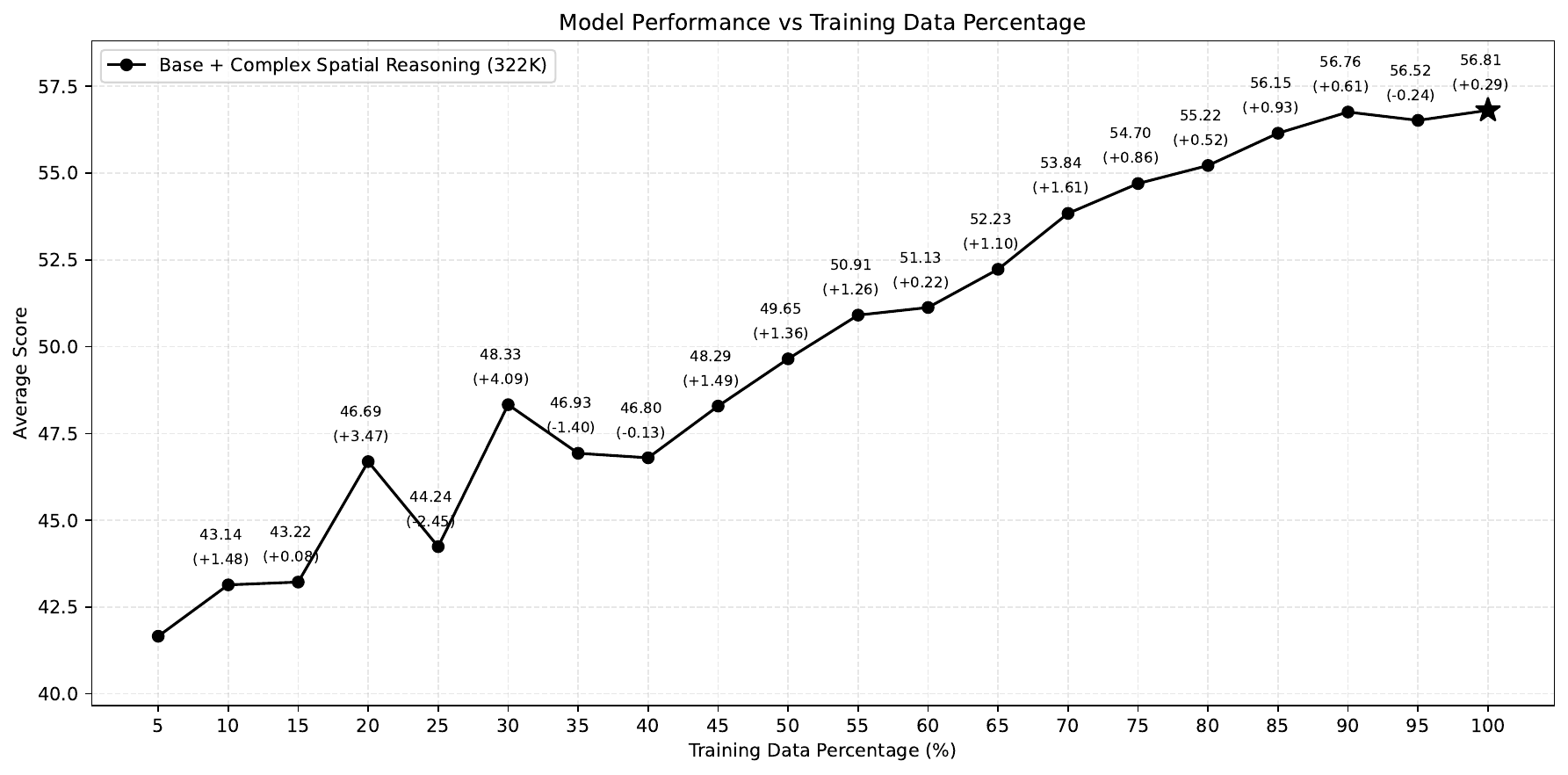}
\caption{
\textbf{Impact of training data size on ViCA2-7B performance on VSI-Bench.}
The average score consistently improves as the percentage of ViCA-322K training data increases, demonstrating that \textbf{ViCA2-7B} benefits from larger-scale spatially grounded instruction tuning and has not yet reached saturation with the current 322K samples.
}
\label{fig:csr_scaling}
\end{figure*}

\subsection{Evaluation Setup}

We evaluate our model on the VSI-Bench benchmark, which comprises 5,130 question-answer pairs derived from 288 egocentric indoor videos. The benchmark encompasses eight tasks categorized into configurational, measurement estimation, and spatiotemporal reasoning.

For consistency and reproducibility, we adhere strictly to the evaluation protocols outlined in the VSI-Bench paper. This includes utilizing the provided prompts and employing greedy decoding with parameters set to \texttt{do\_sample=False} and \texttt{temperature=0}. Our evaluation scripts are publicly available to facilitate replication and further research.

\subsection{Overall Performance on VSI-Bench}

ViCA2-7B achieves state-of-the-art results on VSI-Bench, outperforming both proprietary and open-source models by a large margin(Table~\ref{tab:comparison}). It attains an average score of 56.8, exceeding the best proprietary model, Gemini-1.5 Pro (45.4), by 11.4 points, despite having significantly fewer parameters (7B vs. 72B or more).

Performance gains are consistent across most tasks. Notably, ViCA2-7B surpasses all baselines by +20.1 in Absolute Distance and +17.0 in Room Size. The only category where it underperforms is Relative Direction. This is likely due to limited task-specific supervision: our ViCA-322K dataset focuses on object count, absolute and relative distances, object size, room size, and appearance order, but does not explicitly cover relative direction or route planning. Nevertheless, ViCA2-7B still achieves the highest score on Route Planning (+2.1), suggesting strong cross-task generalization from learned spatial reasoning patterns.

\subsection{Impact of Training Data Size}

To investigate the effect of data scale on model performance, we conduct a controlled experiment by training ViCA2-7B on progressively larger subsets of the ViCA-322K dataset. As shown in Figure~\ref{fig:csr_scaling}, model accuracy consistently improves with increasing data percentage, reaching a peak average score of 56.81 at 100\% of the training data. Notably, performance continues to improve even in the final stages (from 95\% to 100\%), in contrast to the saturation trends observed in our prior work with ViCA, where accuracy plateaued earlier under similar conditions.

We attribute this continued improvement to the increased architectural complexity of ViCA2—particularly the dual-encoder design and token balancing mechanism—which requires greater training data to reach full capacity. While ViCA2-7B remains relatively compact in parameter size, its stronger representational capacity enables it to benefit from additional supervision, especially in spatially grounded reasoning tasks.

These findings suggest that the current training data has not yet saturated the model's learning capacity. ViCA2's architecture retains further performance potential within the 7B-8B model scale, and could benefit from additional high-quality spatial data. This opens avenues for future scaling of spatial instruction datasets to match increasingly capable vision-language architectures.

\section{Conclusion}

In this paper, we addressed the pressing challenge of visuospatial cognition in Multimodal Large Language Models (MLLMs), a crucial yet relatively underexplored area for human-like visual intelligence. While existing MLLMs excel at general visual-language tasks, their fine-grained spatial understanding often remains limited. To bridge this gap, we introduced \textbf{ViCA2}, a novel framework building upon \textbf{LLaVA-OneVision}. \textbf{ViCA2} incorporates a dual vision encoder architecture (\texttt{SigLIP} for semantics, \texttt{Hiera} for spatial structure), augmented by a token ratio control mechanism to balance semantic richness and spatial precision under computational constraints. Furthermore, we developed \textbf{ViCA-322K}, a new large-scale dataset with over 322,000 spatially grounded question-answer pairs for targeted instruction tuning of spatial reasoning.

Our comprehensive experiments on the demanding \textbf{VSI-Bench} benchmark unequivocally demonstrate \textbf{ViCA2}'s superior performance. Despite its compact 7B parameter scale, \textbf{ViCA2-7B} significantly outperformed larger open-source counterparts like \textbf{LLaVA-NeXT-Video-72B} and surpassed leading proprietary models such as \textbf{Gemini-1.5~Pro} in average scores. These outcomes highlight the efficacy of our architectural innovations and targeted training data in fostering robust spatial reasoning without relying on prohibitively large models. Our ablation studies further suggest that \textbf{ViCA2}'s architecture can derive continued benefit from larger spatial instruction datasets, indicating its learning potential is not yet saturated.

These findings underscore the potential of modular architectures and specialized data curation in advancing specific MLLM cognitive abilities, offering a path towards more resource-efficient, capable systems. While \textbf{ViCA2} performs strongly on most \textbf{VSI-Bench} tasks, areas like 'Relative Direction' indicate avenues for refinement, likely through expanded and diversified training data.

Future work will focus on expanding \textbf{ViCA-322K} with more diverse spatial tasks, exploring sophisticated dual-encoder fusion mechanisms, and investigating scalability to larger language models. Applying \textbf{ViCA2} to embodied AI tasks like navigation and robotic interaction also presents a promising direction. By open-sourcing \textbf{ViCA2}, its model weights, codebase, and the \textbf{ViCA-322K} dataset, we aim to catalyze further research in visuospatial cognition, contributing to the broader endeavor of building more spatially intelligent MLLMs.

\section*{Limitations}

Despite its promising results, ViCA2 has certain limitations. Firstly, the ViCA-322K dataset, while extensive, exhibits some coverage gaps; for instance, the model's relatively lower performance on 'Relative Direction' tasks on VSI-Bench can be attributed to insufficient explicit training examples for this specific spatial concept. Secondly, due to computational resource constraints, the Hiera vision encoder backbone was kept frozen during the primary fine-tuning stages (Stages 2 and 3). Fine-tuning Hiera alongside other components might unlock further performance gains by allowing deeper adaptation of its spatial feature extraction capabilities to the language model's needs. Thirdly, the current iteration of ViCA-322K predominantly features indoor scenes, which may limit the model's out-of-the-box generalization to complex outdoor environments or other specialized visual domains without additional domain-specific fine-tuning. Lastly, our current architecture employs a straightforward concatenation for fusing features from the dual encoders; exploring more sophisticated, learnable fusion mechanisms could offer additional improvements in how semantic and spatial information are integrated.

\bibliography{custom}

\appendix

\section{Training Record}

\begin{figure*}[t]
    \centering

    \begin{minipage}{\textwidth}
        \centering
        \includegraphics[width=0.32\textwidth]{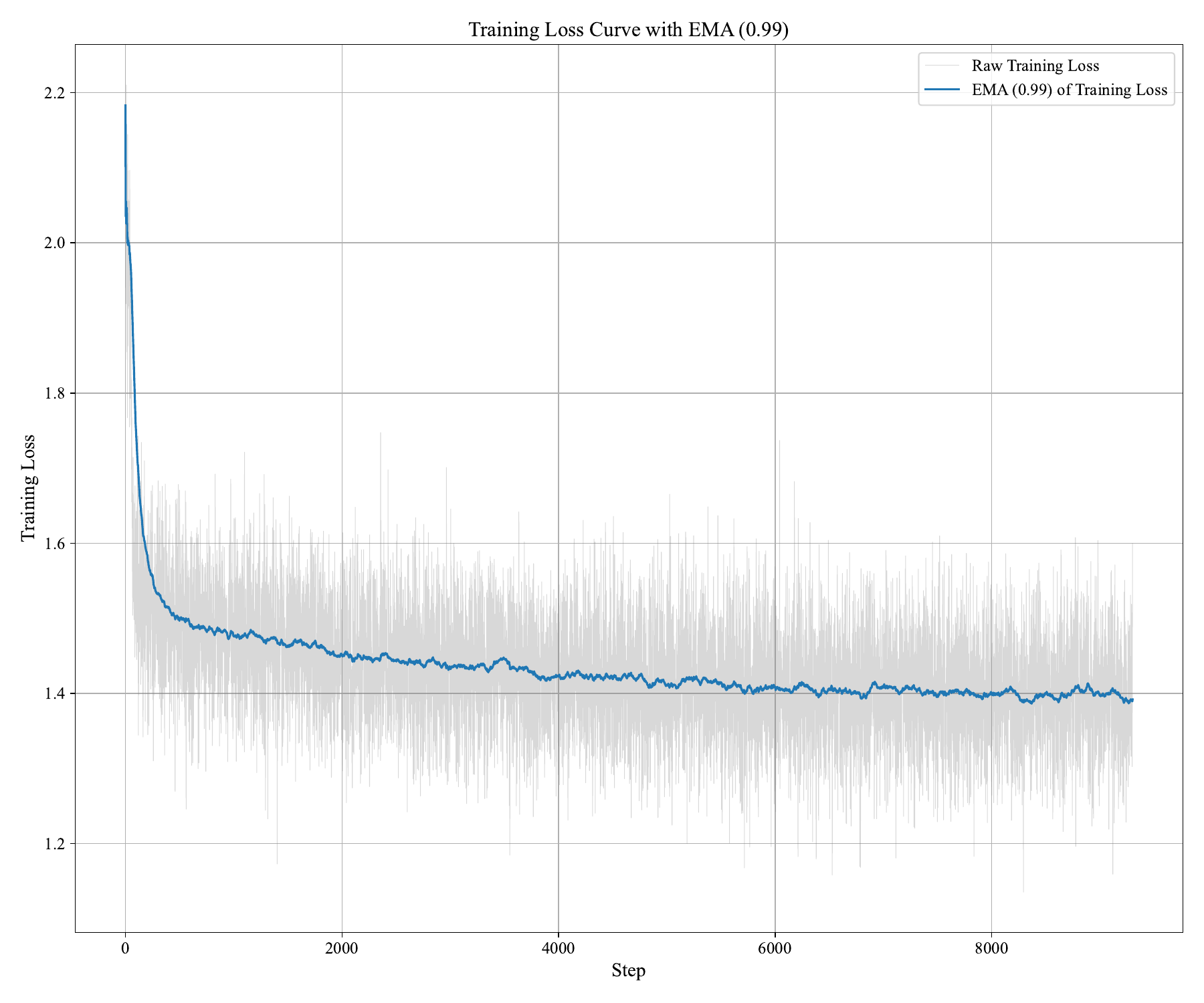}
        \includegraphics[width=0.32\textwidth]{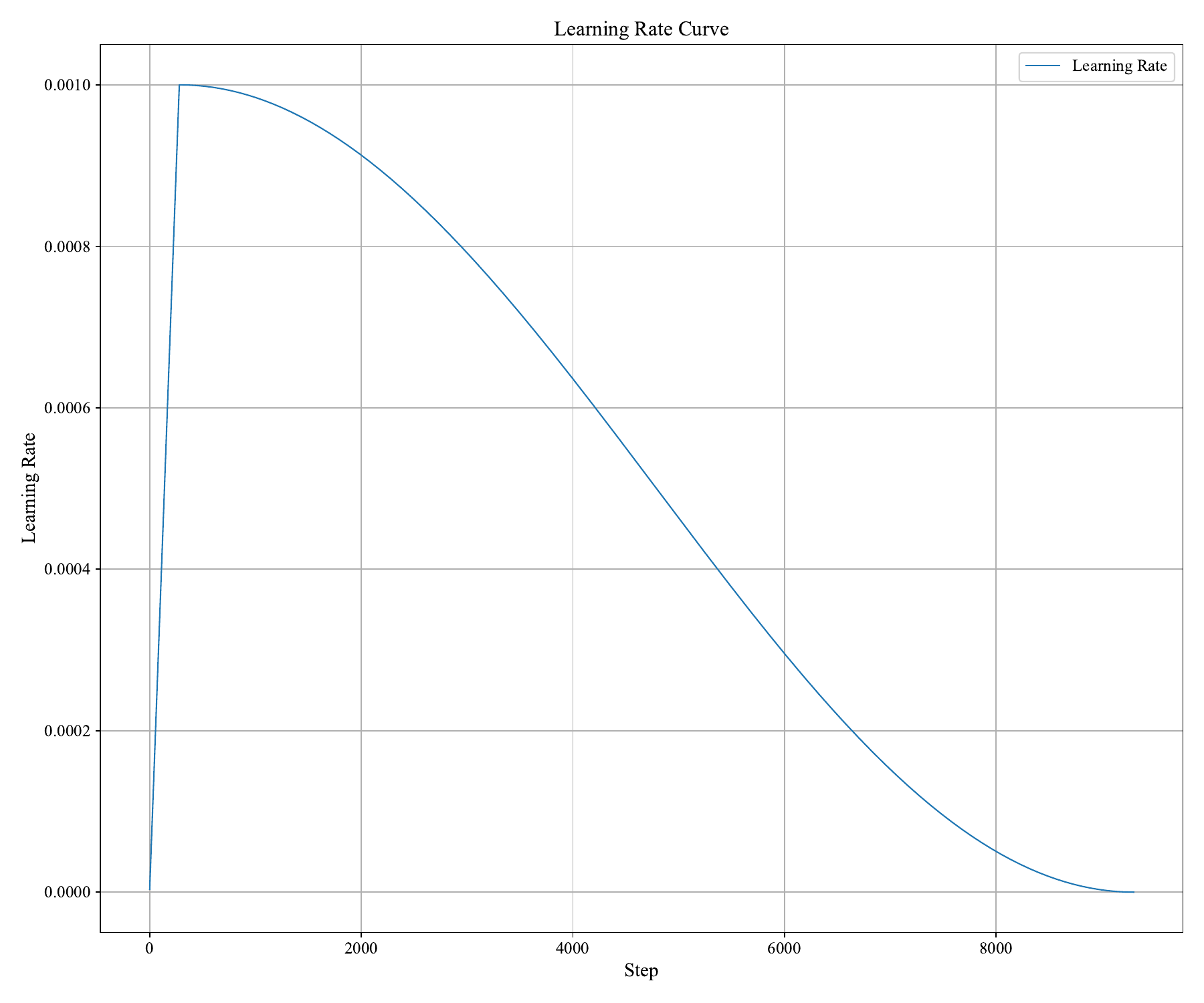}
        \includegraphics[width=0.32\textwidth]{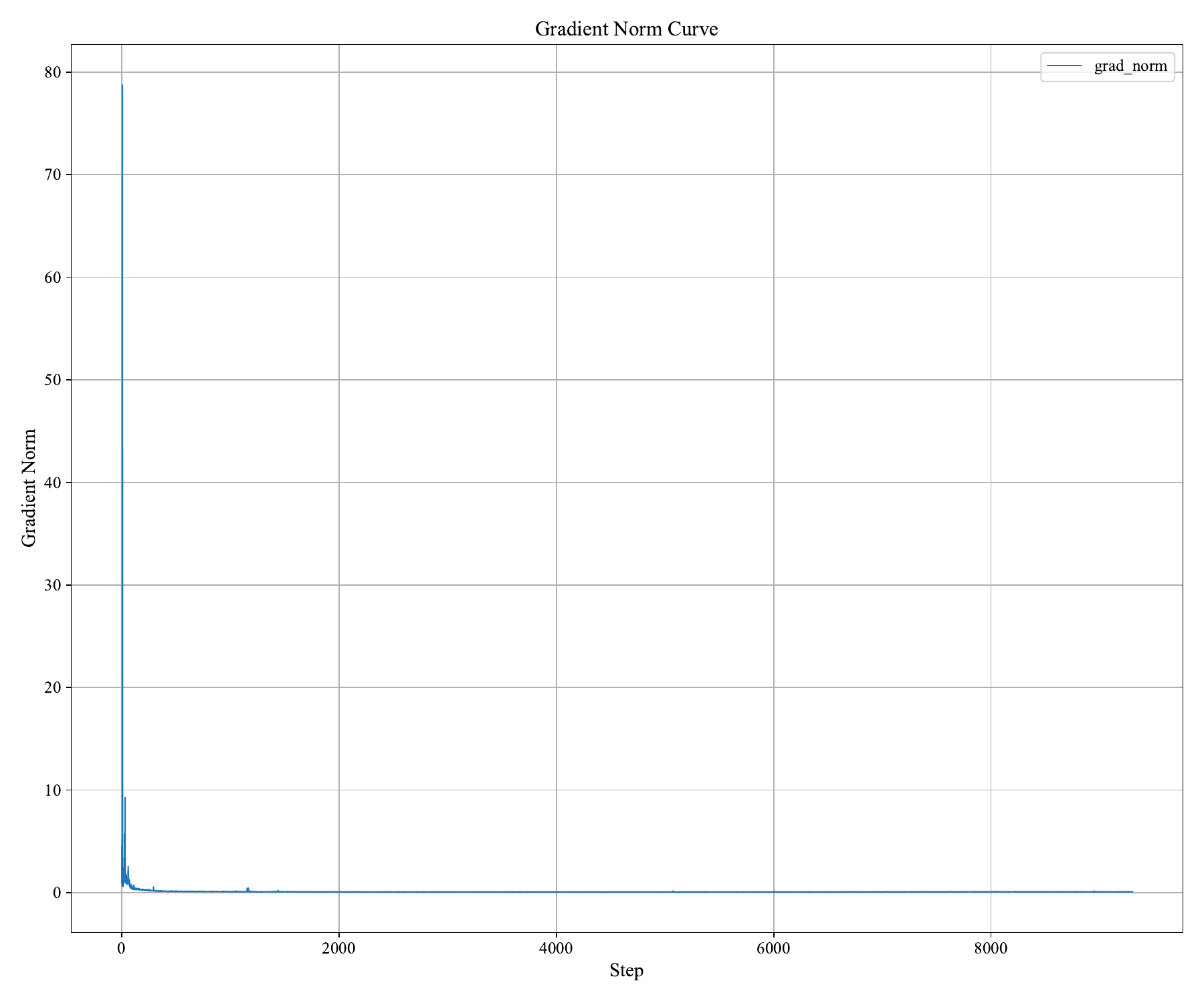}
        \caption*{Stage 1: Training Log}
        \vspace{4mm}
    \end{minipage}

    \begin{minipage}{\textwidth}
        \centering
        \includegraphics[width=0.32\textwidth]{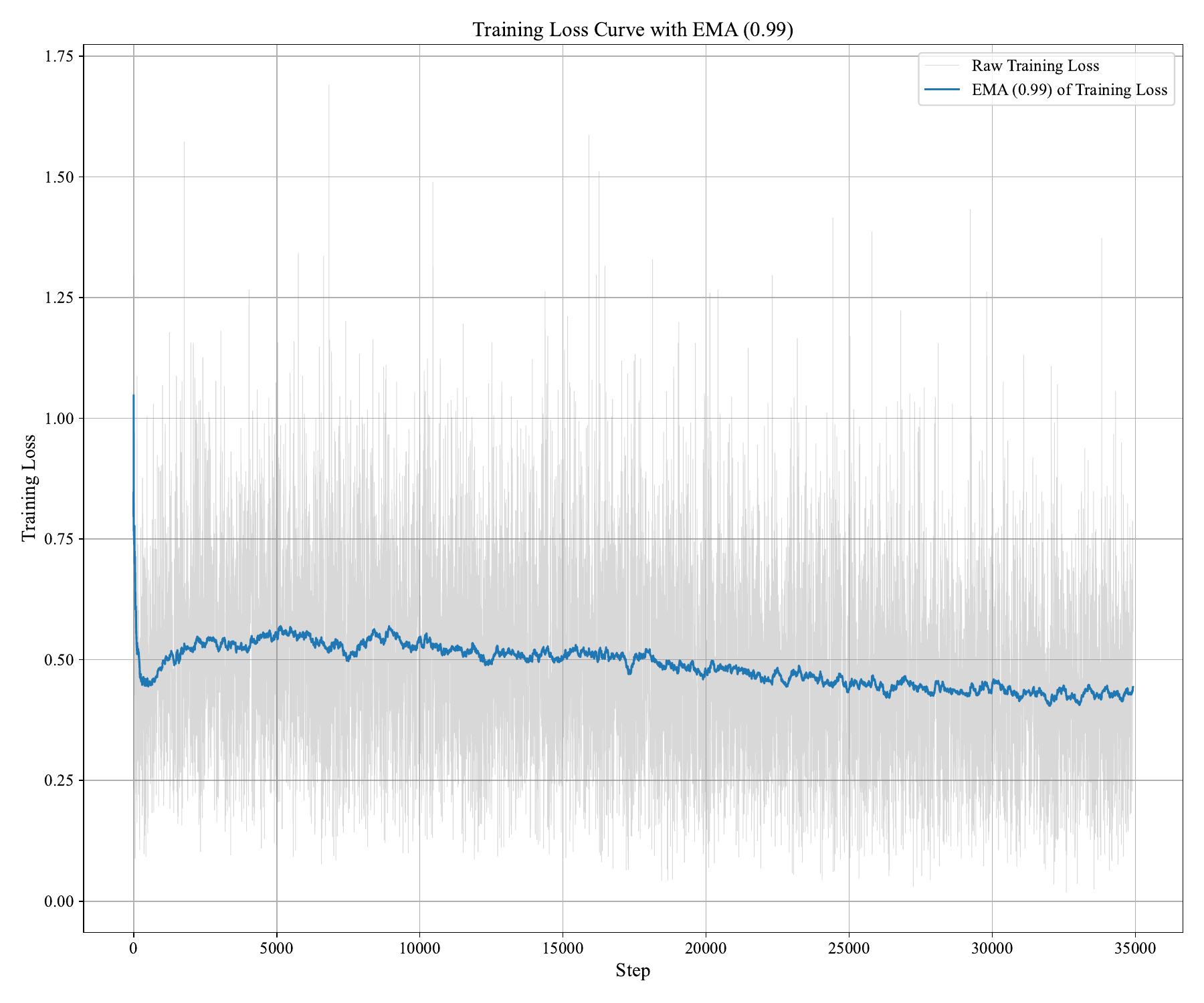}
        \includegraphics[width=0.32\textwidth]{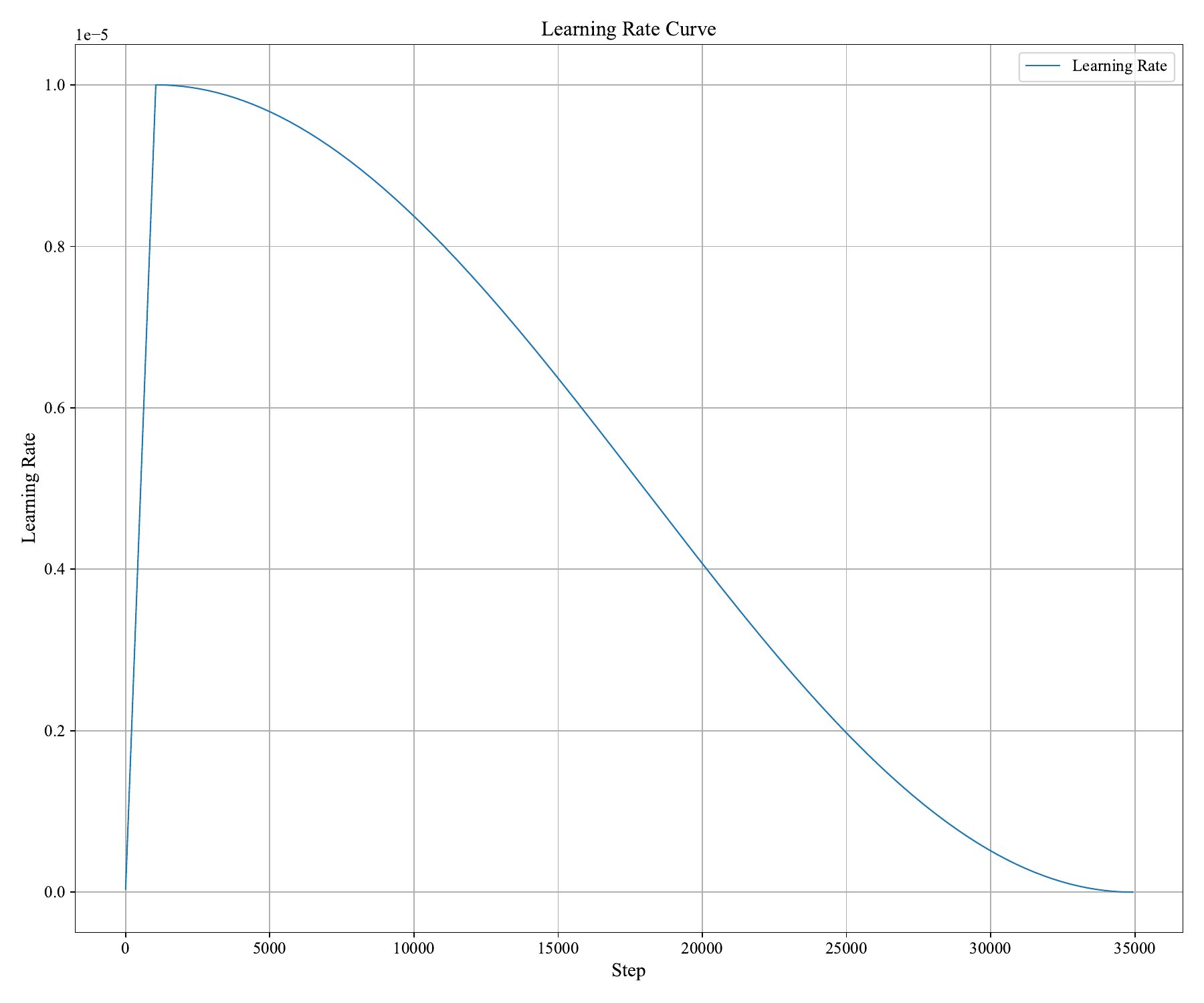}
        \includegraphics[width=0.32\textwidth]{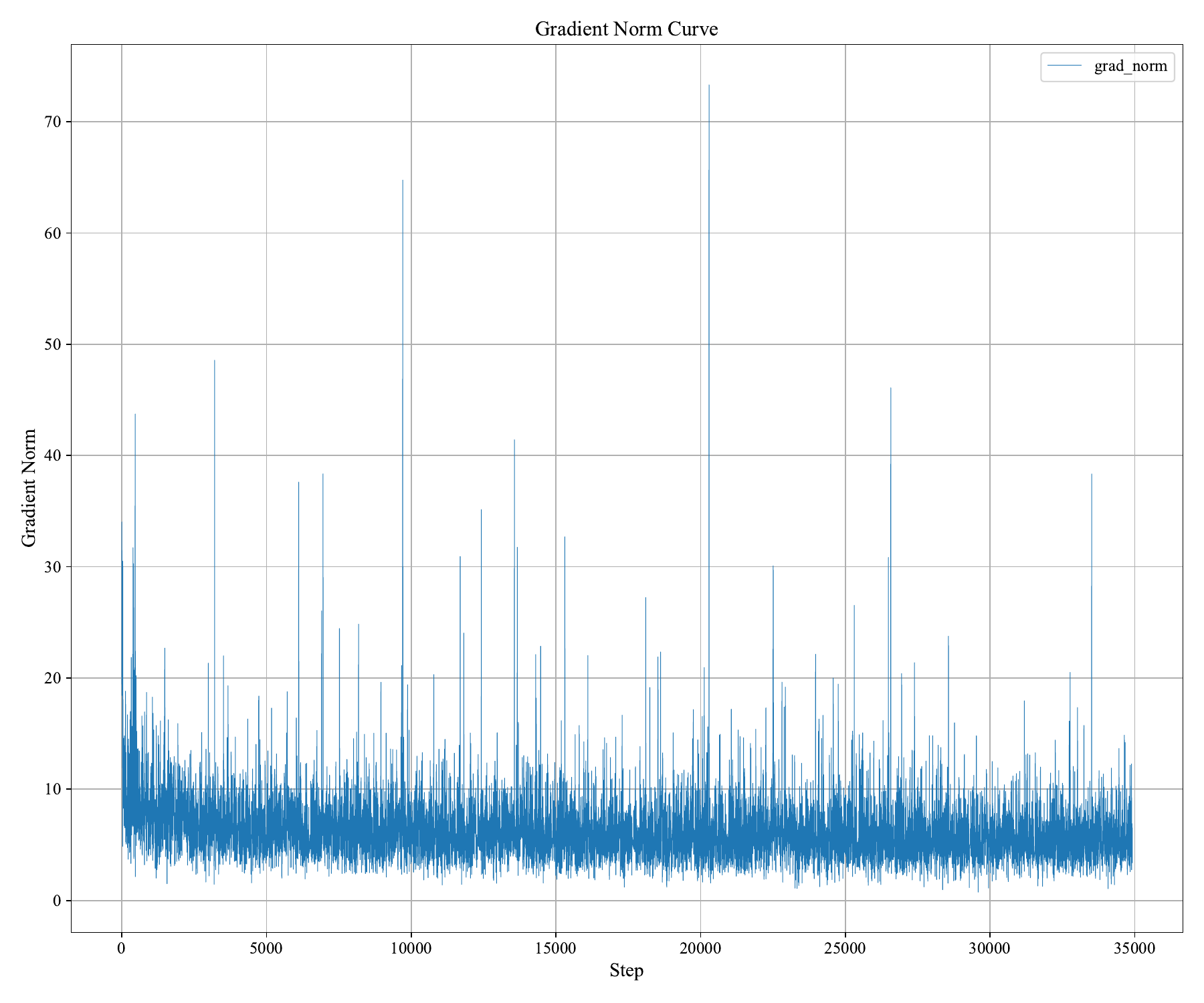}
        \caption*{Stage 2: Training Log}
        \vspace{4mm}
    \end{minipage}

    \begin{minipage}{\textwidth}
        \centering
        \includegraphics[width=0.32\textwidth]{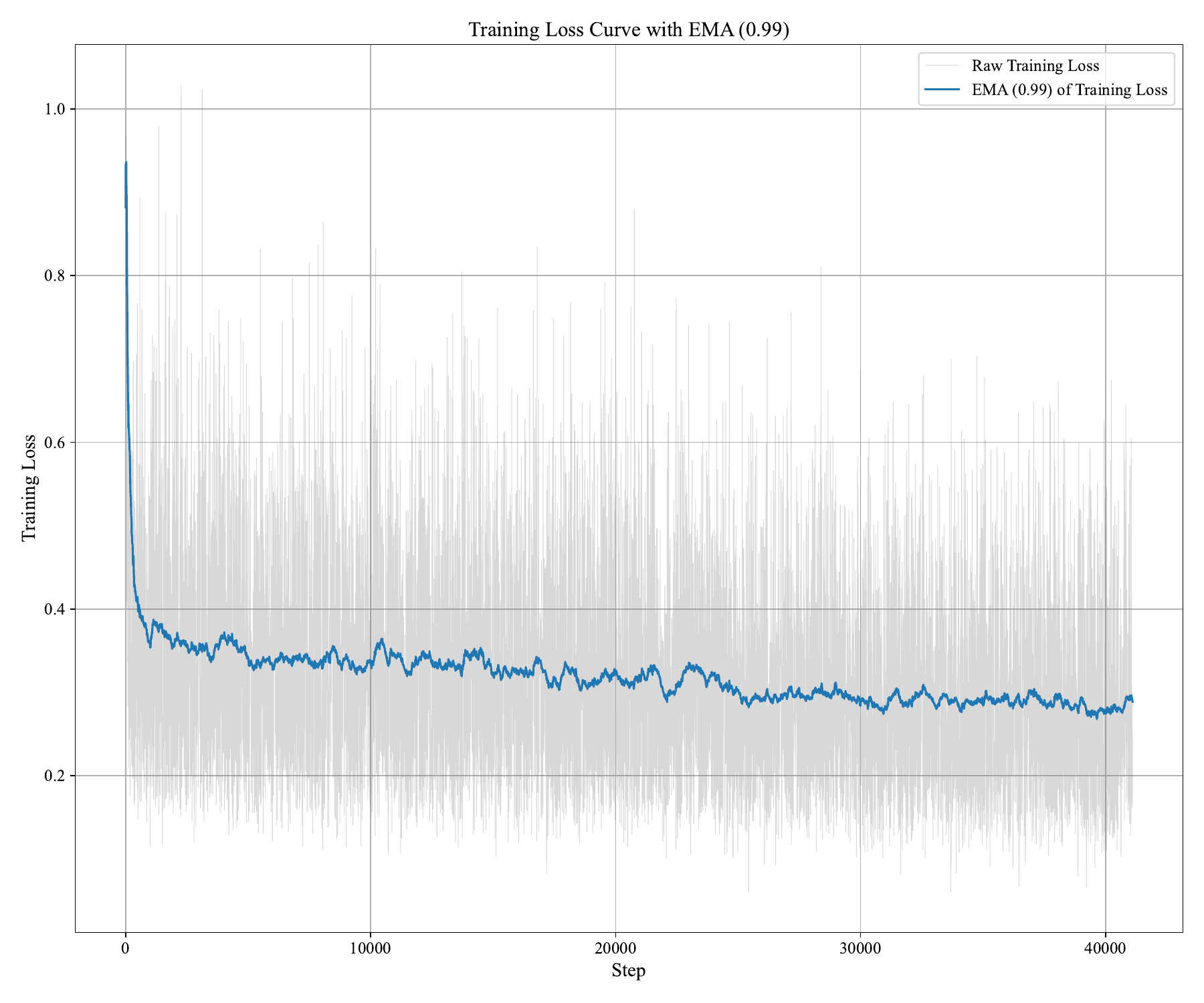}
        \includegraphics[width=0.32\textwidth]{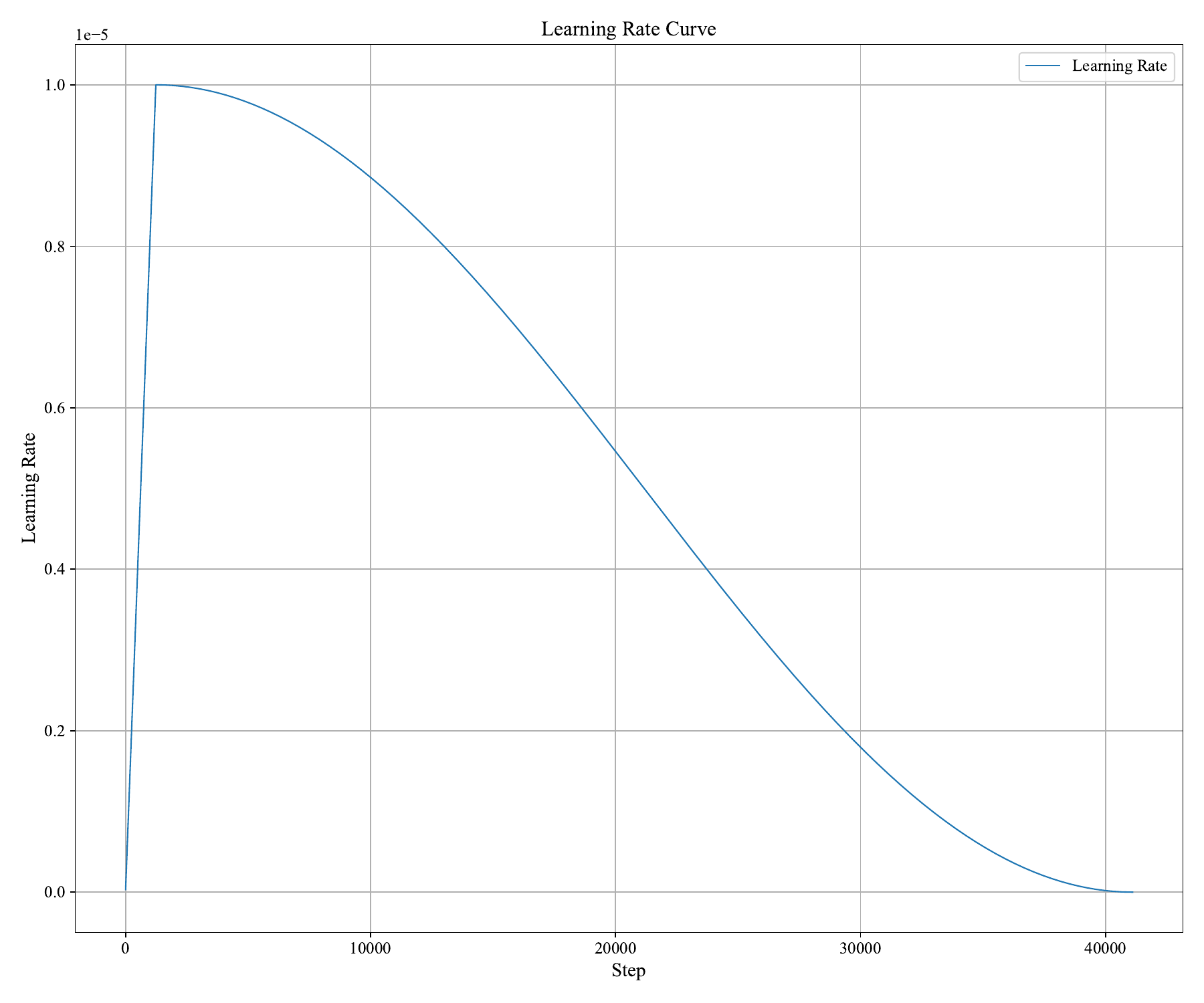}
        \includegraphics[width=0.32\textwidth]{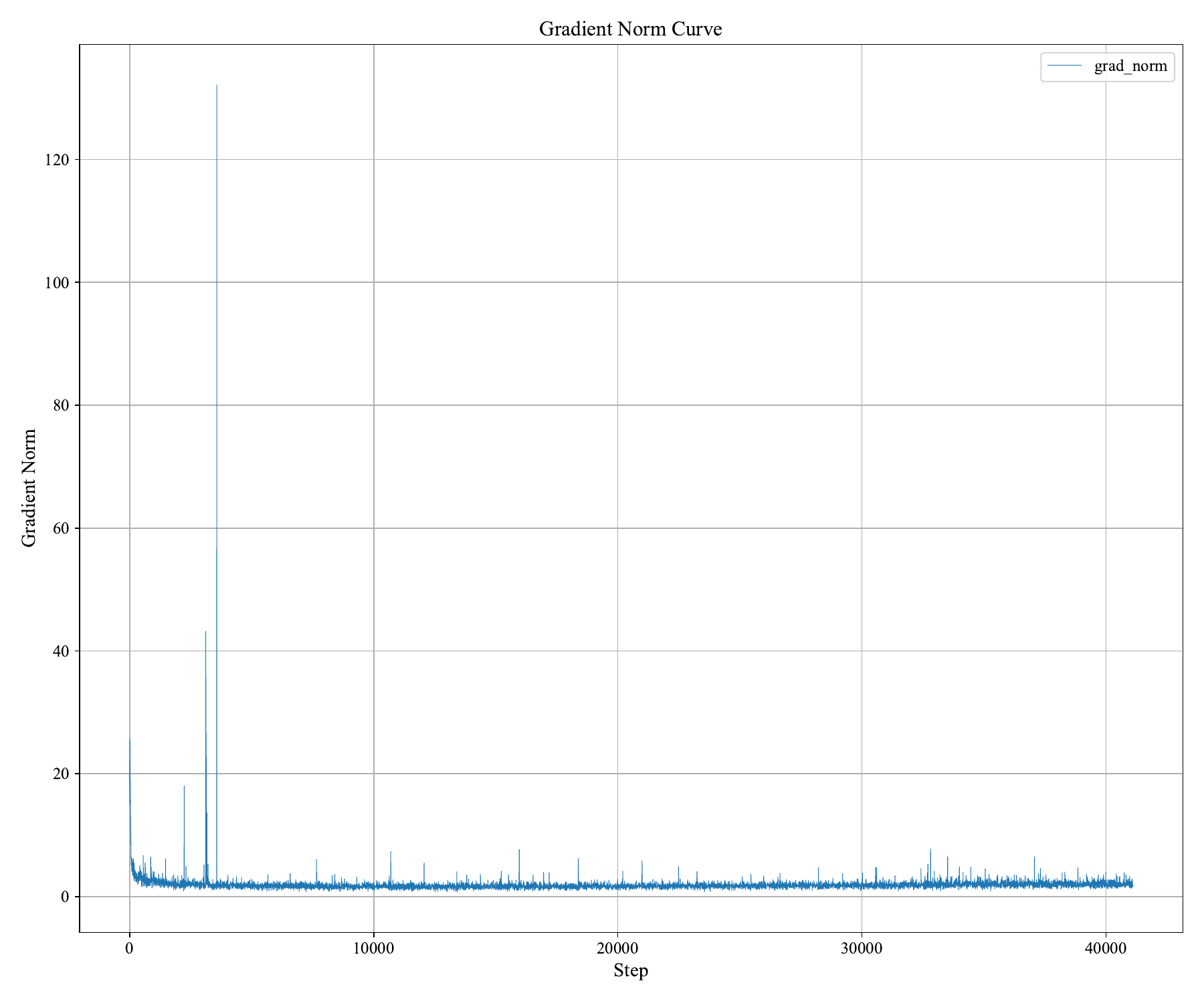}
        \caption*{Stage 3: Training Log}
        \vspace{4mm}
    \end{minipage}

    \begin{minipage}{\textwidth}
        \centering
        \includegraphics[width=0.32\textwidth]{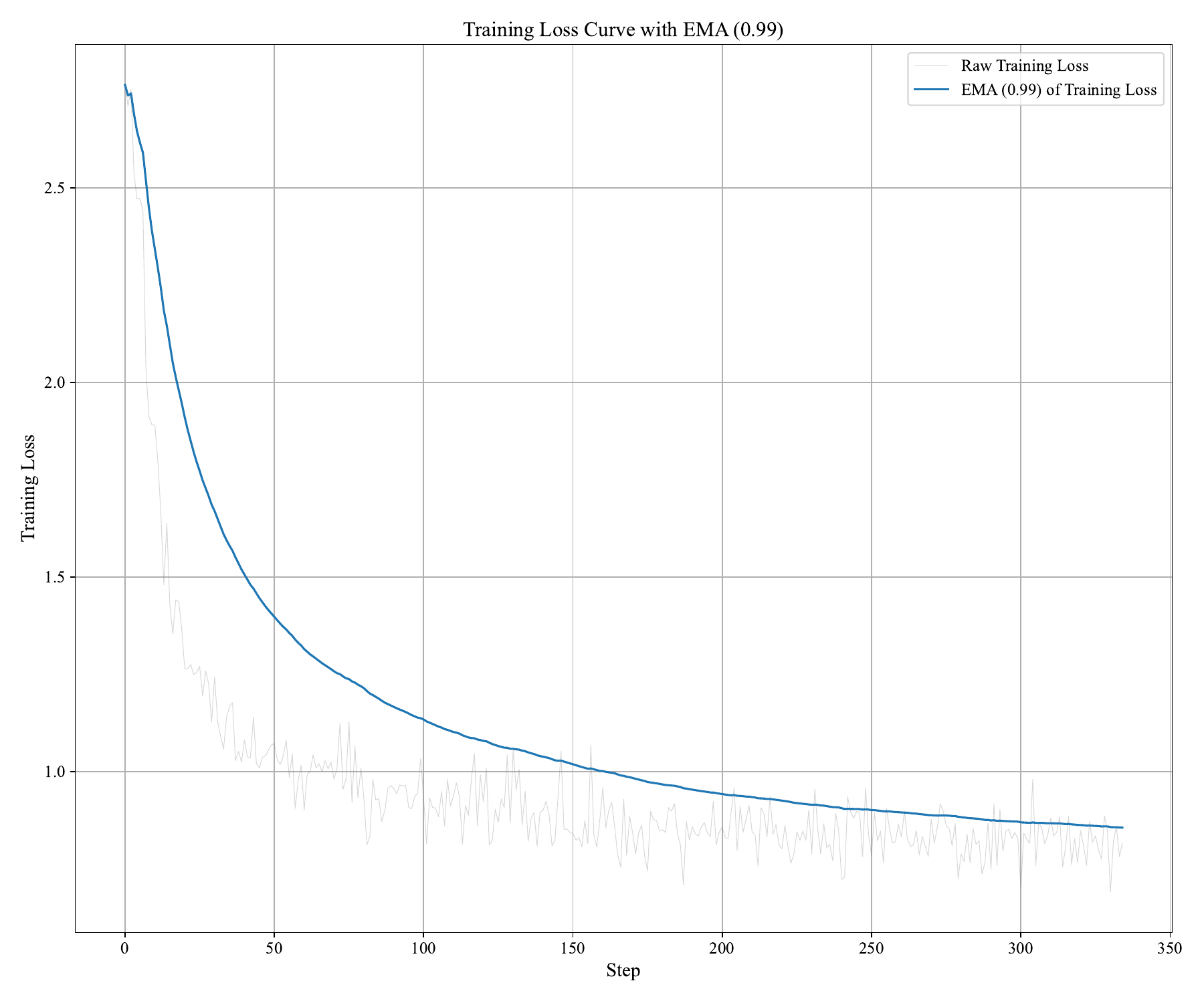}
        \includegraphics[width=0.32\textwidth]{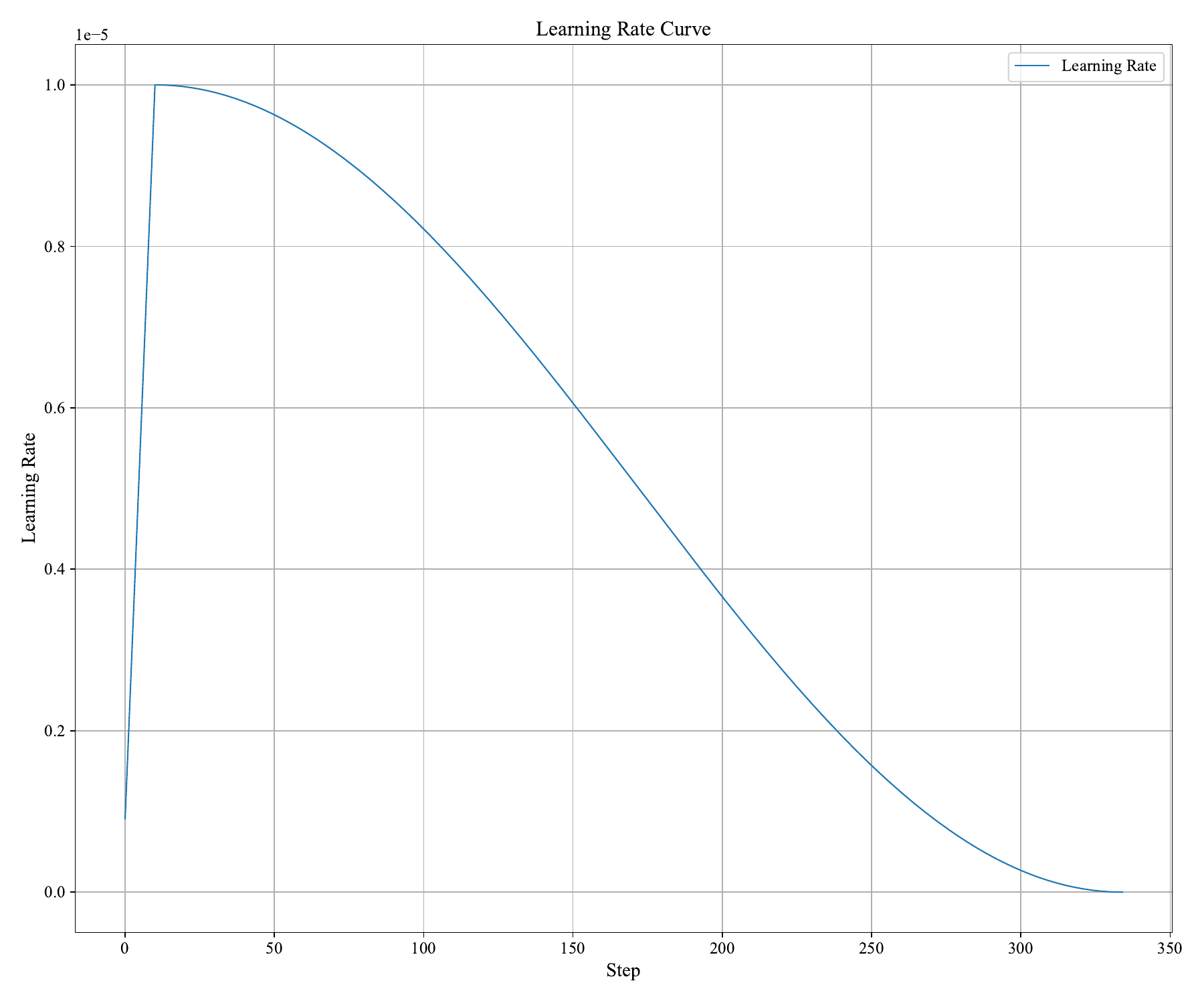}
        \includegraphics[width=0.32\textwidth]{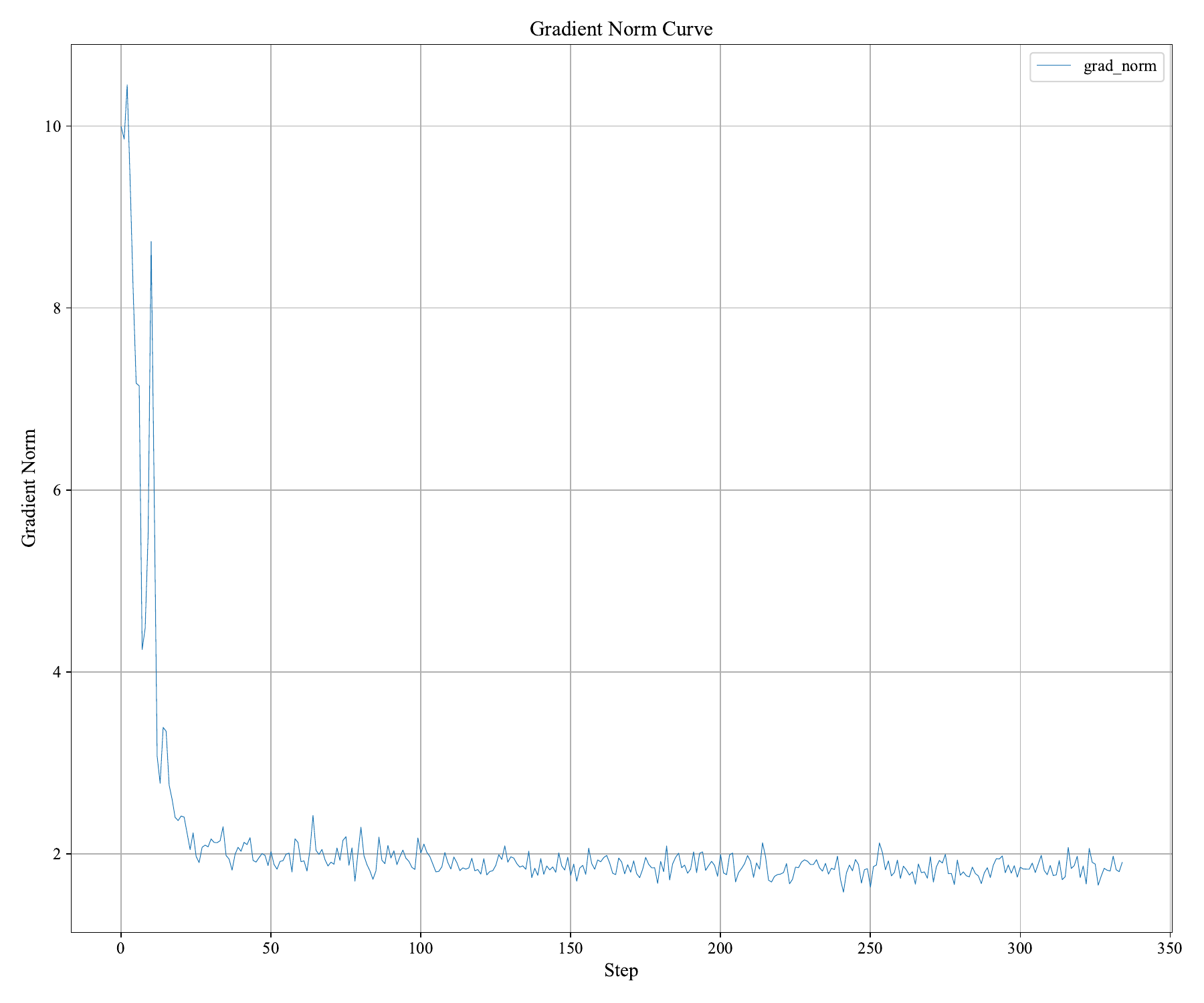}
        \caption*{ViCA2-Thinking: Training Log}
    \end{minipage}

    \caption{\textbf{Training loss curves for the four stages of ViCA2 development.}}
    \label{fig:qualitative_4x3}
\end{figure*}

Figure~\ref{fig:qualitative_4x3} presents the training logs for all four stages of ViCA2. \footnote{For a more detailed and interactive view of these logs, please refer to our public \href{https://wandb.ai/fengqi2016/huggingface/reports/ViCA2-Training-Record--VmlldzoxMjYwMTMzMg?accessToken=rkli8vbwmjs642v5yhq6hk4vs9q1qvrgbotjo6iajq27t0r12s6s8yn0wpesflen}{Weights \& Biases report}.} In both Stage 1 (pretraining the Hiera projection module) and Stage 3 (spatial reasoning fine-tuning), the training loss exhibits a sharp decline during the initial steps, followed by a steady, gradual decrease. This indicates that the model was able to stably learn meaningful patterns from the training data.

In contrast, Stage 2 does not show such a steep drop in loss at the beginning. This is expected, as the base model—initialized from \texttt{lmms-lab/LLaVA-Video-7B-Qwen2}—had already undergone extensive fine-tuning on the same dataset (\texttt{lmms-lab/LLaVA-OneVision-Data}). Nevertheless, due to the introduction of a randomly initialized Hiera projection module, we conducted partial re-tuning on a 10\% subset of the dataset. The goal was to reconcile the representational gap between the Hiera and SigLIP projections and restore the model's overall capability in understanding and describing general videos.

During the final training stage, \textbf{ViCA2-Thinking} (see Appendix~\ref{app:vica2_thinking_details}), the training loss again decreases sharply at the beginning from its initial high value for this complex task. It eventually plateaus around 0.9. This final loss is notably higher than the baseline loss of approximately 0.3 observed for the model (as ViCA2-7B) on simpler, preceding tasks before transitioning to explicit multi-step reasoning. This is likely due to the limited size of the \textbf{ViCA-thinking-2.68K} dataset. Transitioning the model from answering numerical or multiple-choice questions to generating explicit multi-step reasoning processes presents a substantially higher level of complexity.

We believe that the current dataset, while valuable, is insufficient for fully unlocking the model's potential in spatial reasoning with natural language explanations. In future work, we plan to construct a significantly larger and more diverse dataset specifically designed for spatially grounded reasoning, to further advance the visuospatial cognition abilities of multimodal large language models.

\section{Complete Model Architecture}

\begin{figure*}[t]
  \centering
  \includegraphics[width=\textwidth]{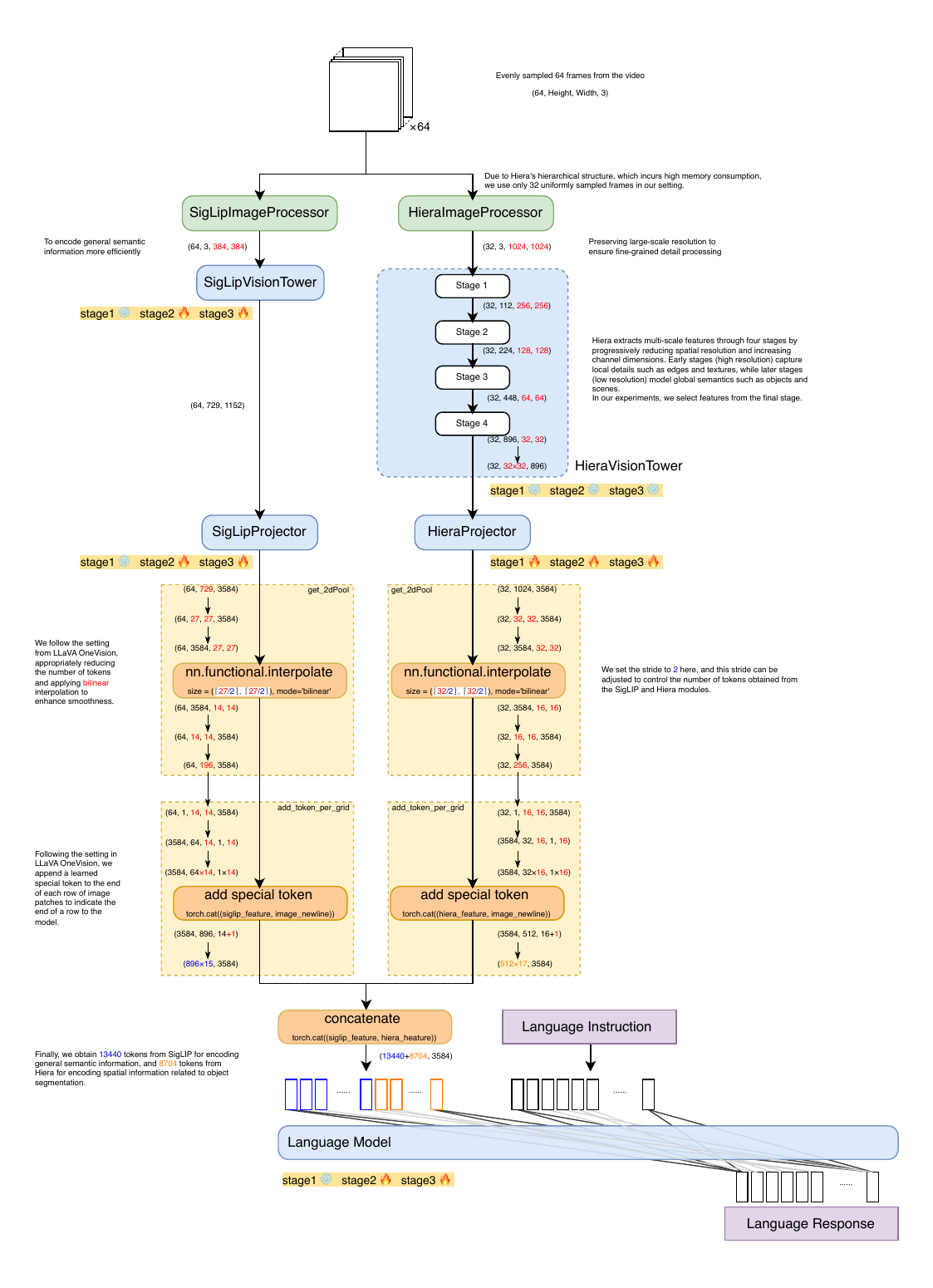} 
  \caption{\textbf{An overview of the ViCA2 architecture.} It integrates \texttt{SigLIP} for global semantics and \texttt{Hiera} for spatial structure, with token ratio control to balance expressiveness and memory efficiency.}
  \label{fig:vica2_arch_complete}
\end{figure*}

Due to space limitations in the main text, a simplified version of the model architecture was provided(Figrue~\ref{fig:vica2_arch}). In this section, we present the full schematic diagram of the complete model architecture(Figrue~\ref{fig:vica2_arch_complete}).

\clearpage

\clearpage

\clearpage

\begin{figure*}[t]
  \centering
  \includegraphics[width=\textwidth]{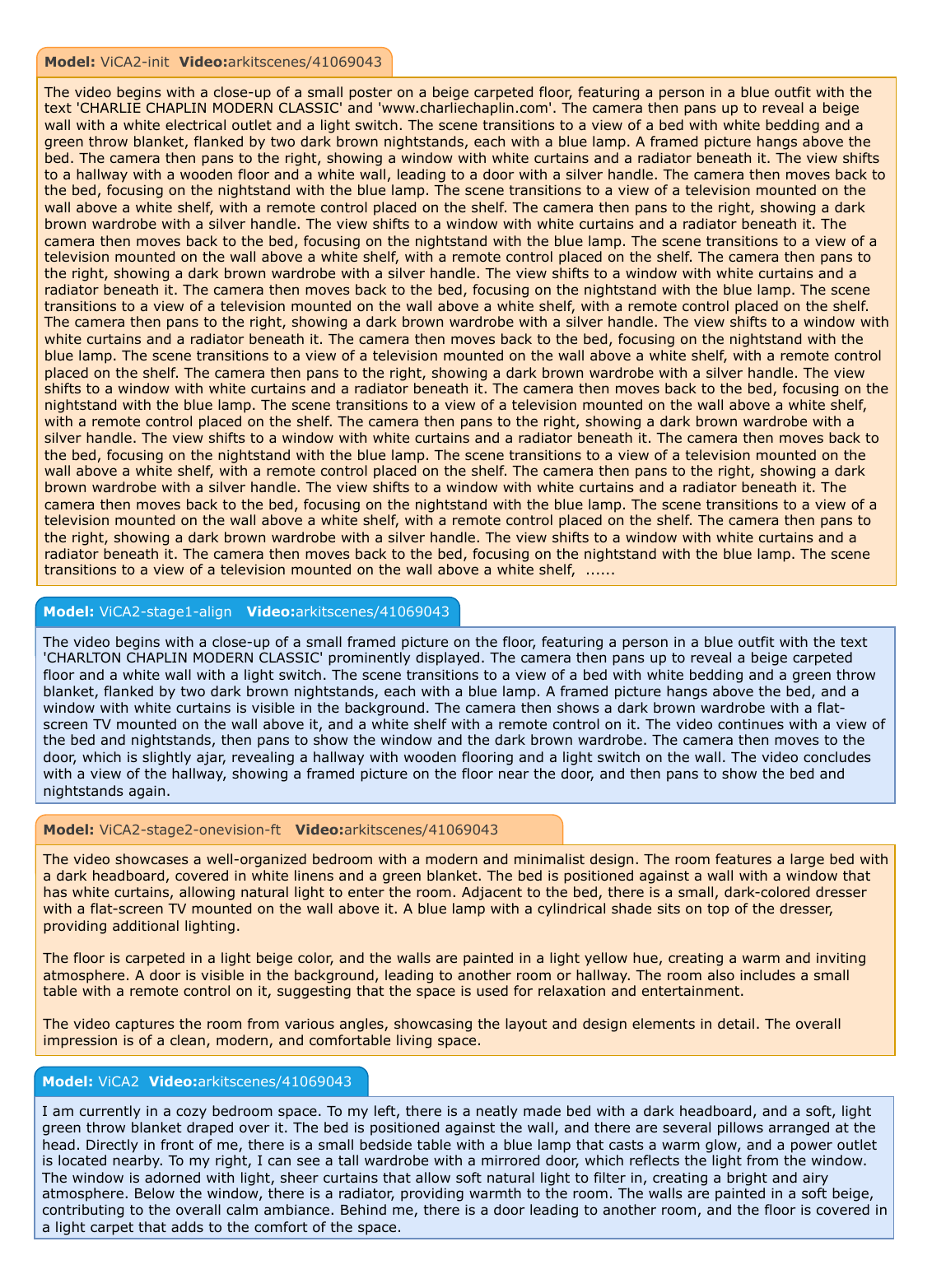} 
  \caption{Qualitative comparison of the generated texts from four different checkpoints on a shared video sample (ARKitScenes/41069043), highlighting the evolution of generation quality during training.}
  \label{fig:qualitative_comparisons_of_the_four_checkpoints}
\end{figure*}

\clearpage

\section{ViCA2-Thinking Fine-Tuning and Observations}
\label{app:vica2_thinking_details} 

\begin{figure*}[t]
  \centering
  \includegraphics[width=\textwidth]{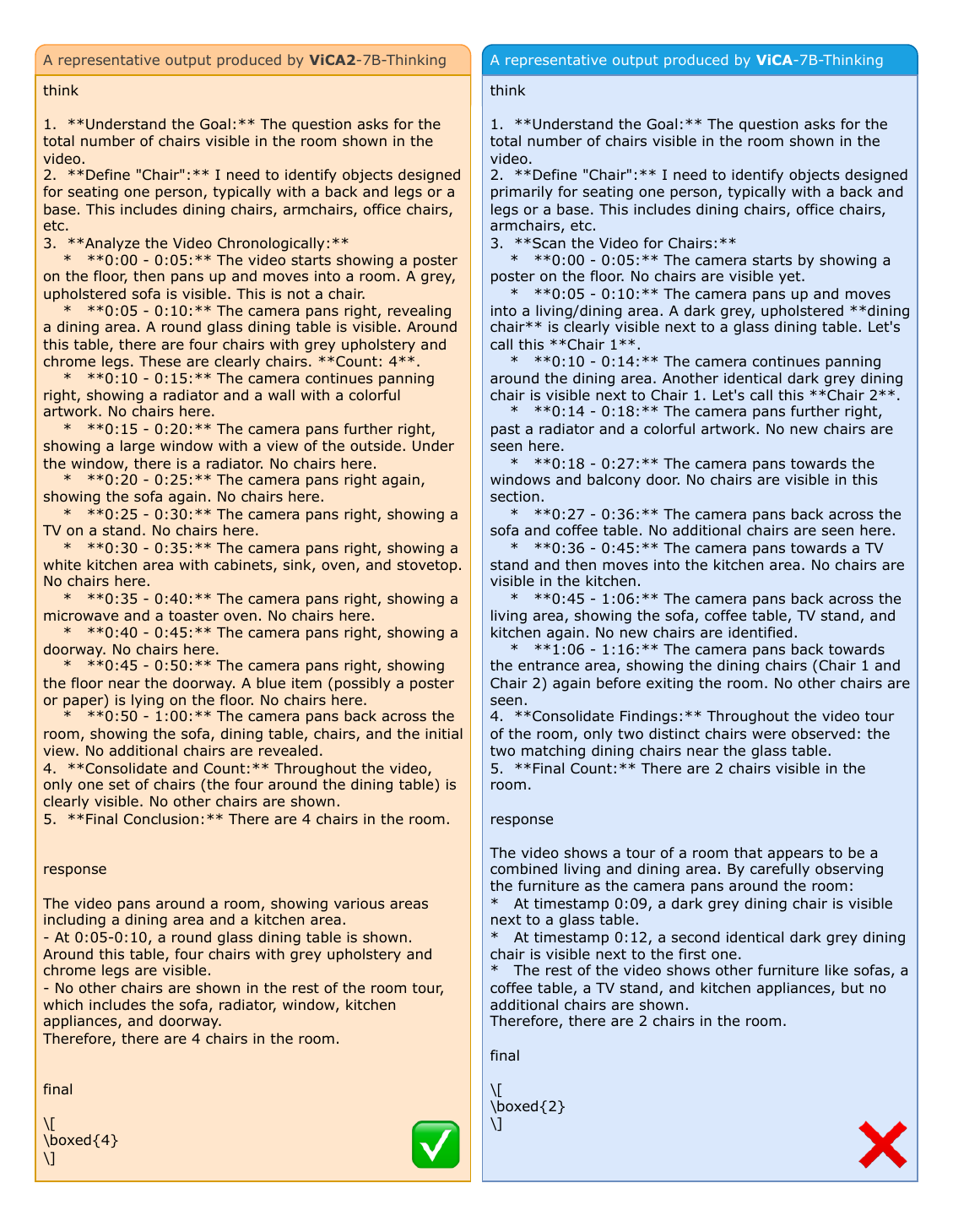}
  \caption{
  Comparison of outputs from \textbf{ViCA2-7B-Thinking} and \textbf{ViCA-7B-Thinking}.
  }
  \label{fig:qualitative1}
\end{figure*}

\begin{figure*}[!t]
  \centering
  \includegraphics[width=\textwidth]{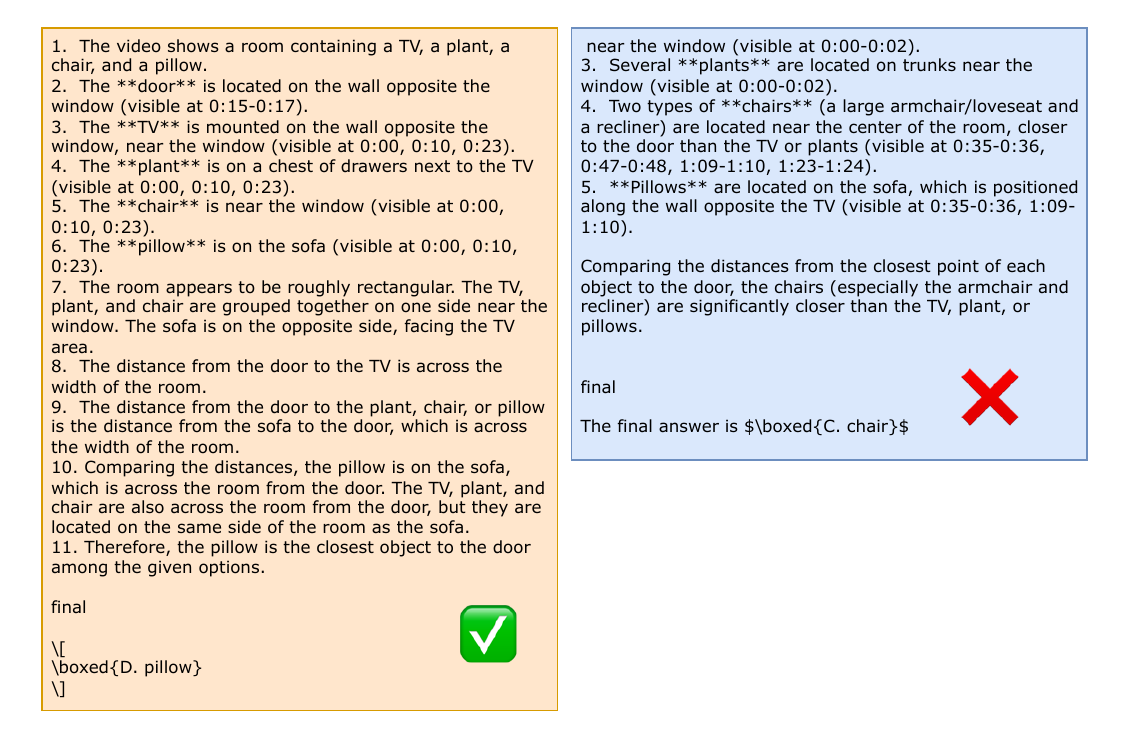}
  \caption{
  Comparison of outputs from \textbf{ViCA2-7B-Thinking} and \textbf{ViCA-7B-Thinking}.
  }
  \label{fig:qualitative2}
\end{figure*}

\begin{figure*}[!t]
  \centering
  \includegraphics[width=\textwidth]{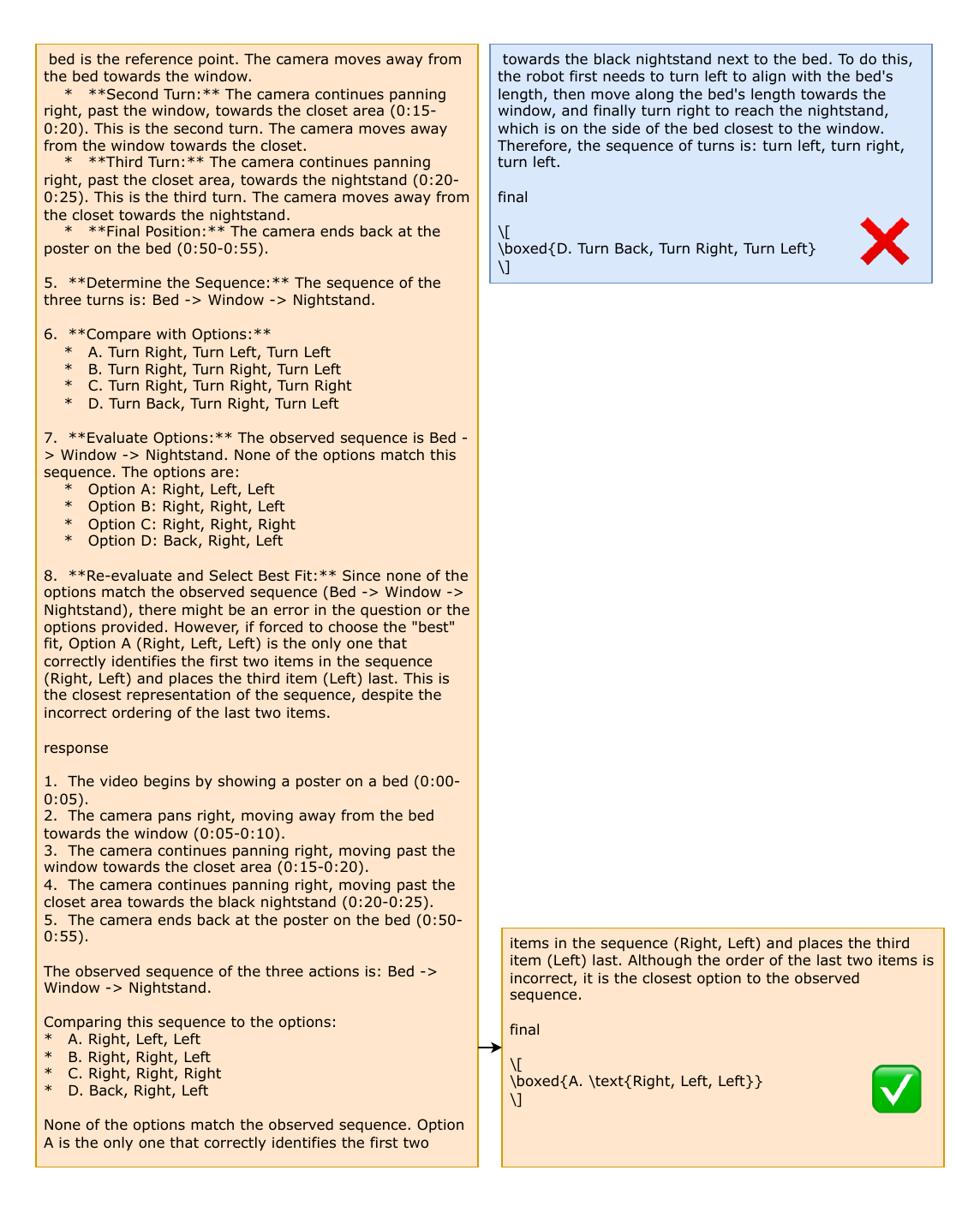}
  \caption{
  Comparison of outputs from \textbf{ViCA2-7B-Thinking} and \textbf{ViCA-7B-Thinking}.
  }
  \label{fig:qualitative3}
\end{figure*}

Following the setup described in our previous ViCA paper, we fine-tuned the ViCA2-7B model using our curated \textbf{ViCA-Thinking-2.68K} dataset. The fine-tuning was conducted on four H200 GPUs (141GB each), and completed within 1 hour and 59 minutes. The resulting model, which we denote as \textbf{ViCA2-7B-Thinking}, is capable of producing an intermediate \textit{Thoughts} segment prior to generating the final response. This reasoning step aims to enhance the model's spatial cognition capabilities. We provide representative output examples from the model in Figure~\ref{fig:qualitative1}, \ref{fig:qualitative2}, \ref{fig:qualitative3}.

In terms of visuospatial performance, ViCA2-7B-Thinking demonstrates a notable drop compared to its predecessor,\textbf{ViCA2-7B}, mirroring the degradation previously observed in \textbf{ViCA-7B-Thinking} (see Table~\ref{tab:comparison_thinking}). Its performance converges to the level of baseline 7B/8B open-source models that have not undergone any spatial reasoning-specific fine-tuning.

We hypothesize that this degradation is linked to the model's ability to perform high-quality text generation. The \textit{Thoughts} component demands significantly more from the model's generative capacity compared to structured QA formats. As we showed in Section~\ref{ssec:stage-wise_results_and_observations}, the model's output fluency and informativeness improve progressively with additional training. However, upon examining the outputs from the checkpoint at the end of Stage 3, we observed that almost all video descriptions begin with highly templated and egocentric phrases such as:

\begin{quote}
\textit{“I find myself in a compact kitchen space...”} 

\textit{“I am seated on a comfortable gray couch...”} 

\textit{“I am currently in a small, functional restroom space...”} 

\textit{“In this kitchen space, I can see a variety of essential appliances...”}
\end{quote}

This observation suggests that the model, during Stage 3 training, acquired a strong tendency to generate egocentric spatial descriptions—likely learned from the spatial caption subset of ViCA-322K. While such descriptions capture local spatial relationships from a first-person perspective, they do not generalize to broader forms of spatial reasoning or flexible narrative generation. Furthermore, the volume of this spatial caption data within ViCA-322K is relatively limited compared to other data types, which restricts the diversity of egocentric descriptive patterns the model can internalize.

We identify three levels of capability relevant to this discussion:

\begin{enumerate}
\item Describing object-level spatial relations from an egocentric (first-person) point of view.
\item Providing general descriptions of the overall spatial layout or environment.
\item Performing free-form, coherent, and abstract natural language generation.
\end{enumerate}

These abilities increase in difficulty from (1) to (3). Our findings suggest that ViCA2-7B, after Stage 3, primarily exhibits improvements in the first capability. This may partially explain why the overall performance of ViCA2-7B-Thinking degrades after fine-tuning with ViCA-Thinking-2.68K, despite its high relevance to spatial reasoning.

\begin{table*}[t]
    \small
    \centering
    \resizebox{\textwidth}{!}{ 
    \begin{tabular}{l|c|cccc|cccc}
    \toprule
    \textbf{Method} & \textbf{Average} & \multicolumn{4}{c|}{\textbf{Numerical Answer}} & \multicolumn{4}{c}{\textbf{Multiple-Choice Answer}} \\
      &  & Obj. Count & Abs. Dist. & Obj. Size & Room Size & Rel. Dist. & Rel. Dir. & Route Plan & Appr. Order \\
    \midrule
    \rowcolor{gray!20} \multicolumn{10}{l}{\textit{Proprietary Models (API)}} \\
    GPT-4o & 34.0 & 46.2 & 5.3 & 43.8 & 38.2 & 37.0 & 41.3 & 31.5 & 28.5 \\
    Gemini-1.5 Flash & 42.1 & 49.8 & 30.8 & 53.5 & 54.4 & 37.7 & 41.0 & 31.5 & 37.8 \\
    Gemini-1.5 Pro & 45.4 & 56.2 & 30.9 & 64.1 & 43.6 & 51.3 & \cellcolor{gray!40}46.3 & 36.0 & 34.6 \\
    \midrule
    \rowcolor{gray!20} \multicolumn{10}{l}{\textit{Open-source Models-7B/8B}} \\
    InternVL2-8B & 34.6 & 23.1 & 28.7 & 48.2 & 39.8 & 36.7 & 30.7 & 29.9 & 39.6 \\
    VILA-1.5-8B & 28.9 & 17.4 & 21.8 & 50.3 & 18.8 & 32.1 & 34.8 & 31.0 & 24.8 \\
    LLaVA-NeXT-Video-7B & 35.6 & 48.5 & 14.0 & 47.8 & 24.2 & 43.5 & \textbf{42.4} & 34.0 & 30.6 \\
    LLaVA-OneVision-7B & 32.4 & 47.7 & 20.2 & 47.4 & 12.3 & 42.5 & 35.2 & 29.4 & 24.4 \\
    \rowcolor{gray!20} \multicolumn{10}{l}{\textit{ViCA2-7B-thinking}} \\
    ViCA2-7B-thinking (\textbf{ours}) & 35.5 & 33.9 & 25.0 & 42.0 & 33.1 & 40.8 & 34.9 & 33.1 & 41.6 \\
    \midrule
    \rowcolor{gray!20} \multicolumn{10}{l}{\textit{Open-source Models-40B}} \\
    InternVL2-40B & 36.0 & 34.9 & 26.9 & 46.5 & 31.8 & 42.1 & 32.2 & 34.0 & 39.6 \\
    VILA-1.5-40B & 31.2 & 22.4 & 24.8 & 48.7 & 22.7 & 40.5 & 25.7 & 31.5 & 32.9 \\
    \midrule
    \rowcolor{gray!20} \multicolumn{10}{l}{\textit{Open-source Models-72B}} \\
    LLaVA-NeXT-Video-72B & 40.9 & 48.9 & 22.8 & 57.4 & 35.3 & 42.4 & 36.7 & 35.0 & 48.6 \\
    LLaVA-OneVision-72B & 40.2 & 43.5 & 23.9 & 57.6 & 37.5 & 42.5 & 39.9 & 32.5 & 44.6 \\
    \midrule
    \rowcolor{gray!20} \multicolumn{10}{l}{\textit{ViCA2-7B}} \\
    ViCA2-7B (\textbf{ours}) & \cellcolor{gray!40}\textbf{56.8(+11.4)} & \cellcolor{gray!40}\textbf{65.7(+9.5)} & \cellcolor{gray!40}\textbf{51.0(+20.1)} & \cellcolor{gray!40}\textbf{75.5(+11.4)} & \cellcolor{gray!40}\textbf{71.4(+17.0)} & \cellcolor{gray!40}\textbf{51.6(+0.3)} & 34.6 & \cellcolor{gray!40}\textbf{38.1(+2.1)} & \cellcolor{gray!40}\textbf{66.5(+17.9)} \\
    \bottomrule
    \end{tabular}
    }
    \caption{\textbf{Quantitative comparison of ViCA2-7B-Thinking, ViCA2-7B, and selected baseline models on the VSI-Bench benchmark.} While ViCA2-7B achieves the highest average score, ViCA2-7B-Thinking trades off numerical accuracy for improved interpretability through intermediate reasoning steps.}

    \label{tab:comparison_thinking}
    \end{table*}

We further argue that achieving a balanced improvement in both spatial cognition and general language generation requires a revised training strategy. Due to computational constraints, we initialized our model from \texttt{lmms-lab/LLaVA-Video-7B-Qwen2} and fine-tuned it sequentially on a subset of \texttt{LLaVA-OneVision-Data} and then on ViCA-322K. This strategy, while effective for spatial QA tasks with structured outputs (e.g., numeric or multiple-choice answers), may have inadvertently weakened the model's open-ended generation capabilities—an ability that is essential for high-quality reasoning in ViCA2-7B-Thinking.

Nevertheless, our overall training approach did yield strong gains on VSI-Bench, confirming the effectiveness of both our data curation and architectural innovations. We hypothesize that with access to greater compute, a joint training regime combining \texttt{LLaVA-OneVision-Data} and ViCA-322K from the outset—rather than in staged phases—could simultaneously preserve text generation quality and enhance spatial reasoning. Once such a robust base model is obtained, fine-tuning with ViCA-Thinking-2.68K could lead to even greater gains, leveraging the explicit reasoning format introduced by the \textit{Thought} prompts.

We leave the empirical validation of this hypothesis for future work.

\begin{figure*}[t]
  \centering
  \includegraphics[width=\textwidth]{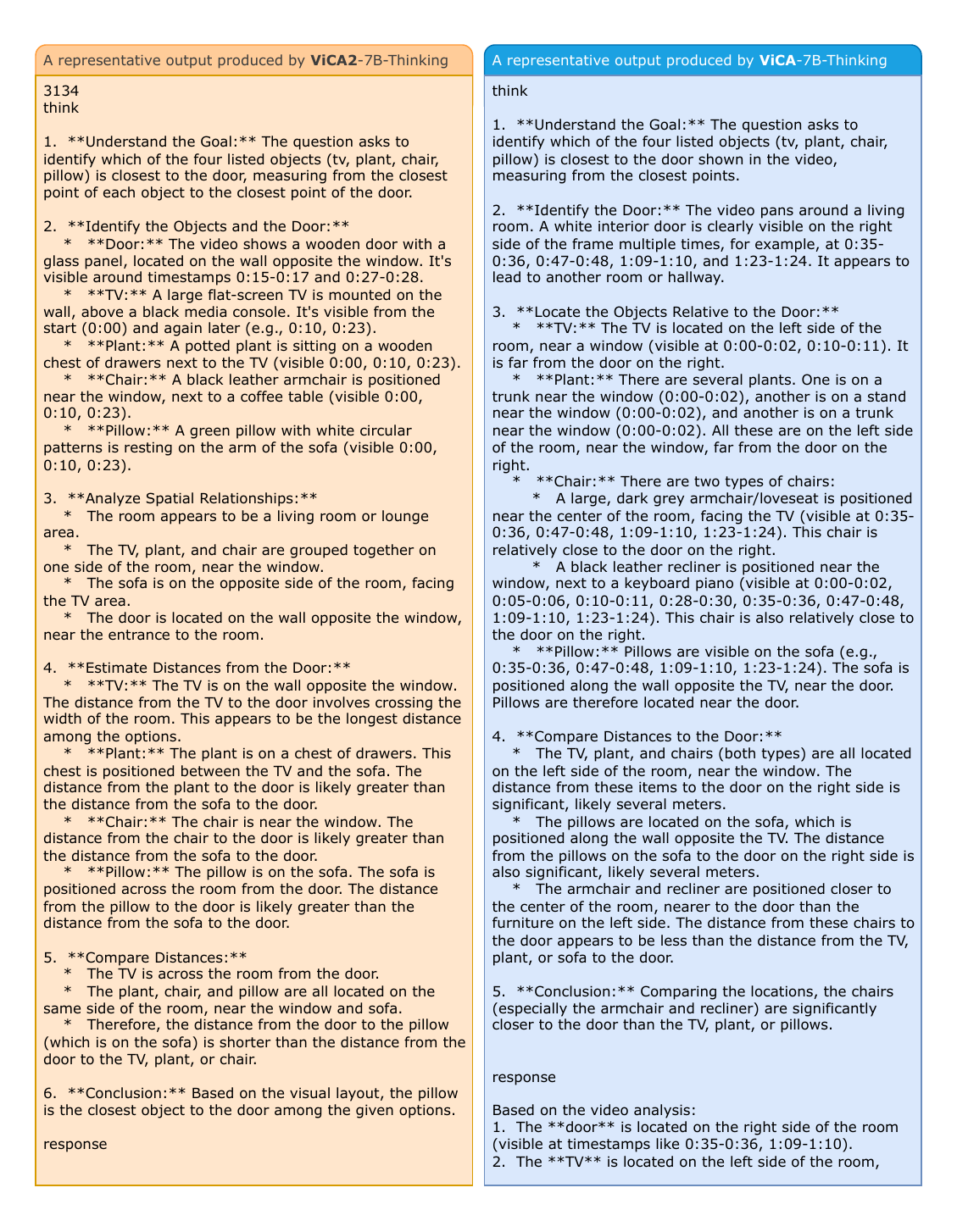}
\end{figure*}

\begin{figure*}[t]
  \centering
  \includegraphics[width=\textwidth]{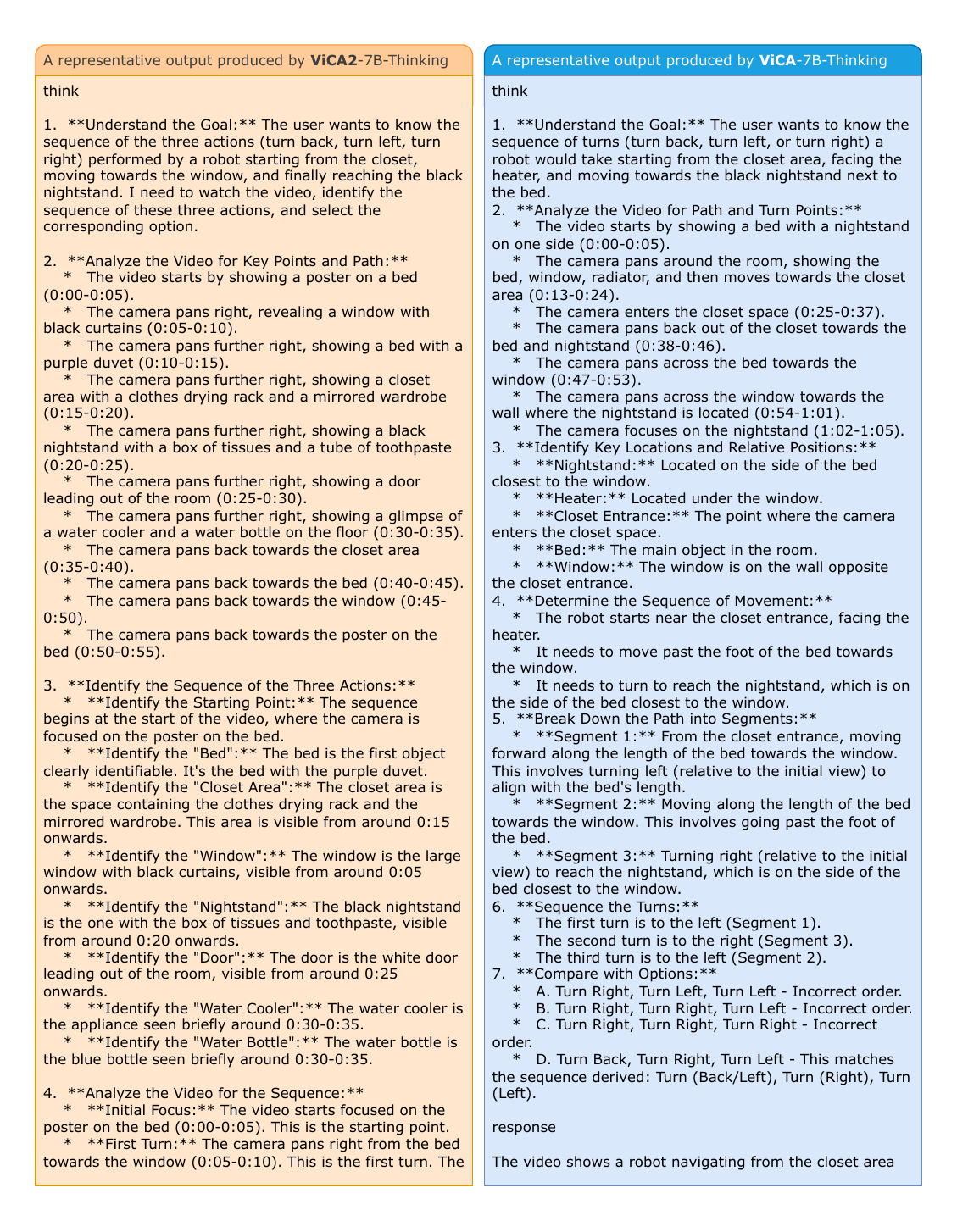}
\end{figure*}

\section{Comparison with ViCA-7B}

While our proposed model, \textbf{ViCA2-7B}, achieves state-of-the-art performance in visuospatial reasoning compared to existing multimodal large language models (see Table~\ref{tab:comparison}), its overall performance is slightly lower than that of our earlier model, \textbf{ViCA-7B}, trained in previous work (see Table~\ref{tab:vica_1_2_comparison}).

\begin{figure*}[t]
  \centering
  \includegraphics[width=\textwidth]{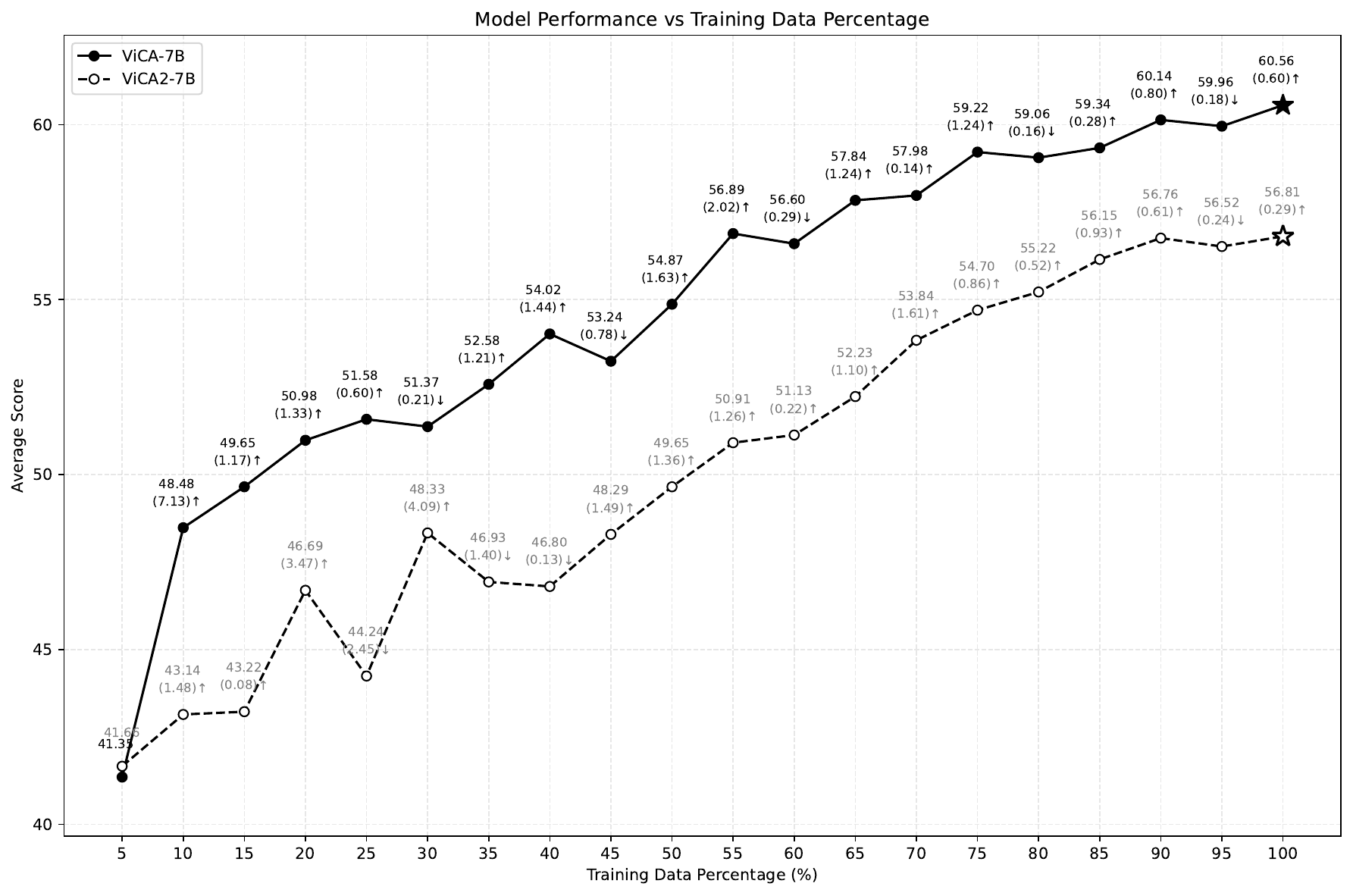}
  \caption{
  Performance Comparison of ViCA and ViCA2 with Increasing Training Data Size
  }
  \label{fig:scaling_comparison}
\end{figure*}

We hypothesize that this discrepancy arises from the increased complexity and capacity of ViCA2-7B. Compared to ViCA-7B, ViCA2-7B incorporates a dual vision encoder architecture and significantly more trainable parameters. Consequently, it likely demands a larger and more diverse dataset to reach its full potential. However, the \textbf{ViCA-322K} dataset—while already extensive—was originally designed and curated for ViCA-7B. During the construction of this dataset, we conducted performance scaling analysis on ViCA-7B and observed that model performance plateaued and slightly declined beyond a certain data scale, prompting us to halt further data collection at that point (Figure~\ref{fig:scaling_comparison}).

\begin{table}[t]
  \small
  \centering
  \setlength{\tabcolsep}{4pt}
  \begin{tabular}{lcc}
    \toprule
    & \textbf{ViCA-7B} & \textbf{ViCA2-7B} \\
    \midrule
    \rowcolor{gray!20} \multicolumn{3}{l}{\textit{Numerical Answer}} \\
    Obj. Count         & \textbf{68.8} & 67.7 \\
    Abs. Dist.         & \textbf{57.0} & 51.0 \\
    Obj. Size          & \textbf{79.2} & 75.5 \\
    Room Size          & \textbf{75.1} & 71.4 \\
    \midrule
    \rowcolor{gray!20} \multicolumn{3}{l}{\textit{Multiple-Choice Answer}} \\
    Rel. Dist.         & \textbf{58.5} & 51.6 \\
    Rel. Dir.          & \textbf{42.6} & 34.6 \\
    Route Plan         & 34.5 & \textbf{38.1} \\
    Appr. Order        & \textbf{68.8} & 66.5 \\
    \midrule
    \textbf{Average}   & \textbf{60.6} & 56.8 \\
    \bottomrule
  \end{tabular}
  \caption{
  Comparison of ViCA-7B and ViCA2-7B Performance on VSI-Bench
  }
  \label{tab:vica_1_2_comparison}
  \end{table}
  
\begin{figure*}[t]
  \centering
  \includegraphics[width=\textwidth]{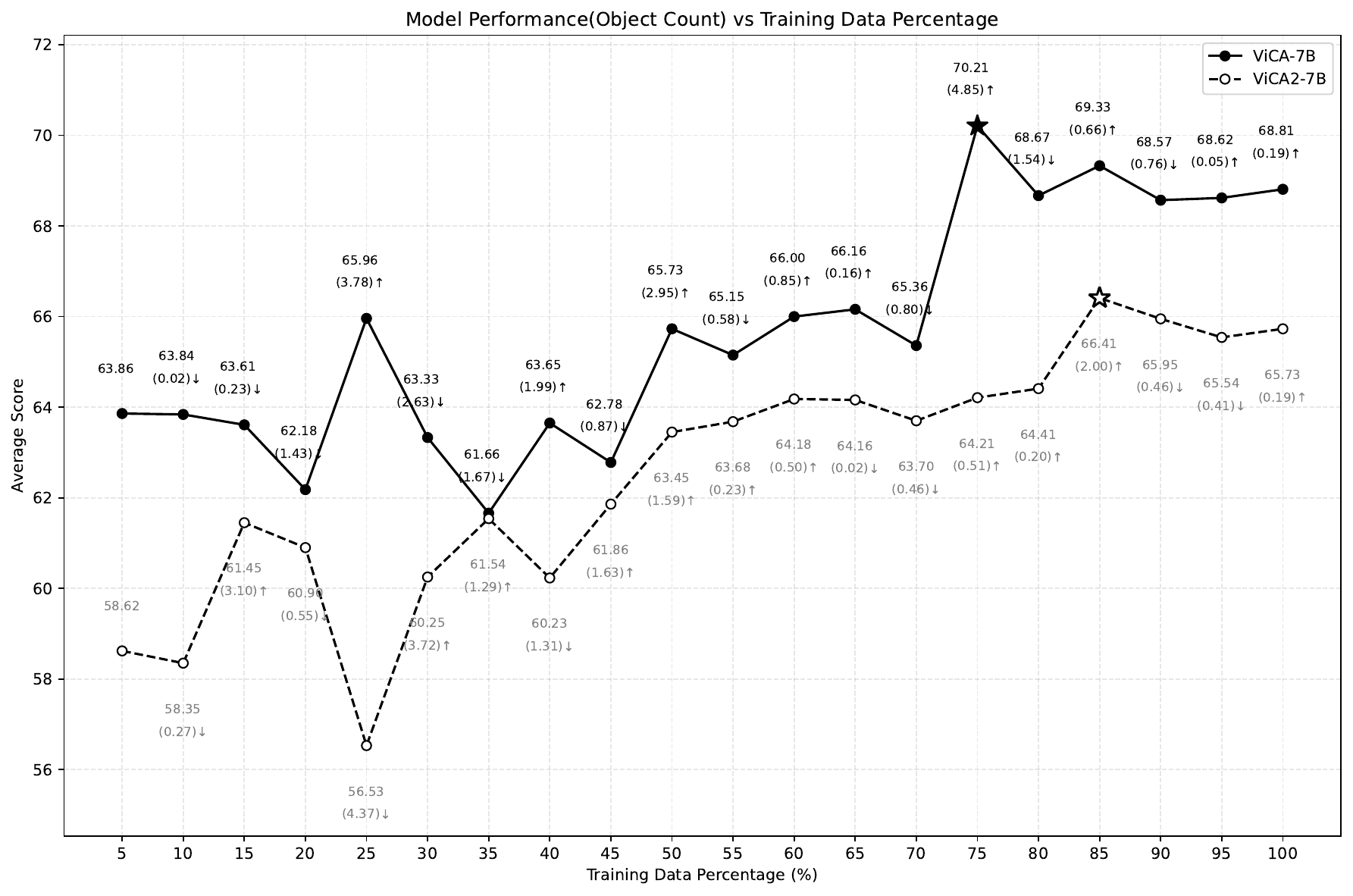}
  \caption{
  Performance Comparison of ViCA and ViCA2 with Increasing Training Data Size - \textbf{Object Count}.
  }
  \label{fig:scaling_comparison1}
\end{figure*}

\begin{figure*}[t]
  \centering
  \includegraphics[width=\textwidth]{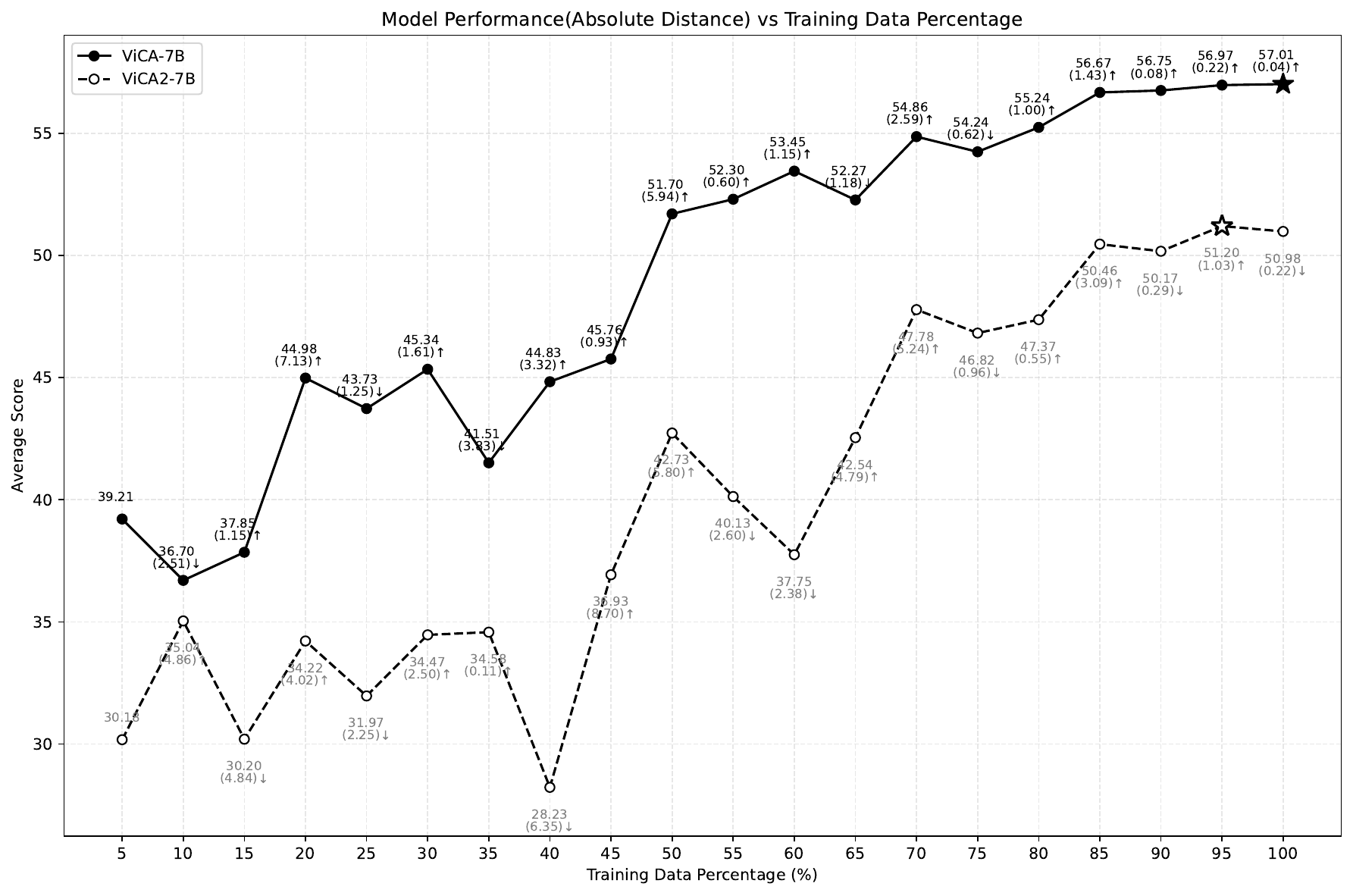}
  \caption{
  Performance Comparison of ViCA and ViCA2 with Increasing Training Data Size - \textbf{Absoulte Distance}.
  }
  \label{fig:scaling_comparison2}
\end{figure*}

\begin{figure*}[t]
  \centering
  \includegraphics[width=\textwidth]{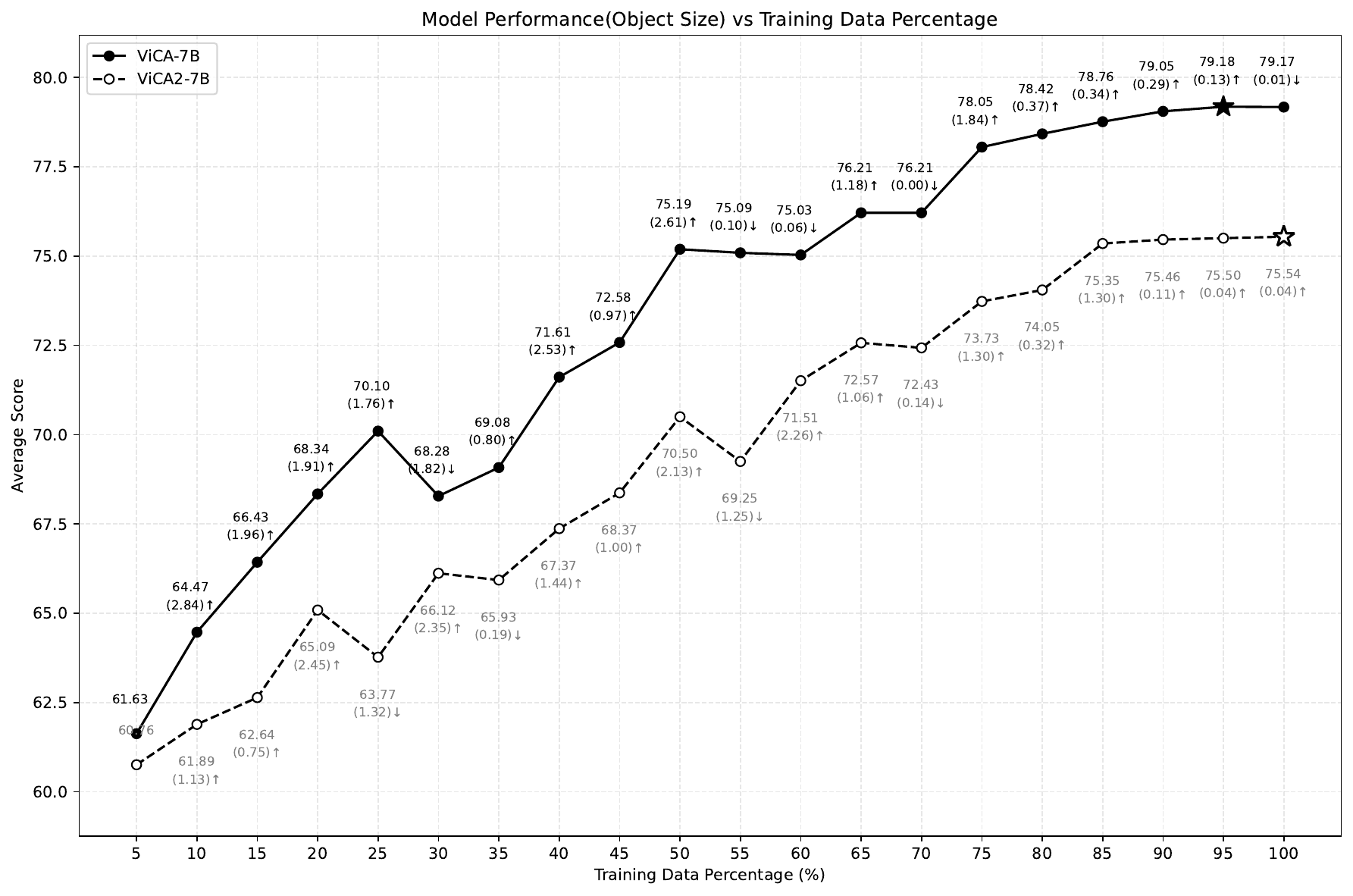}
  \caption{
  Performance Comparison of ViCA and ViCA2 with Increasing Training Data Size - \textbf{Object Size}.
  }
  \label{fig:scaling_comparison3}
\end{figure*}

\begin{figure*}[t]
  \centering
  \includegraphics[width=\textwidth]{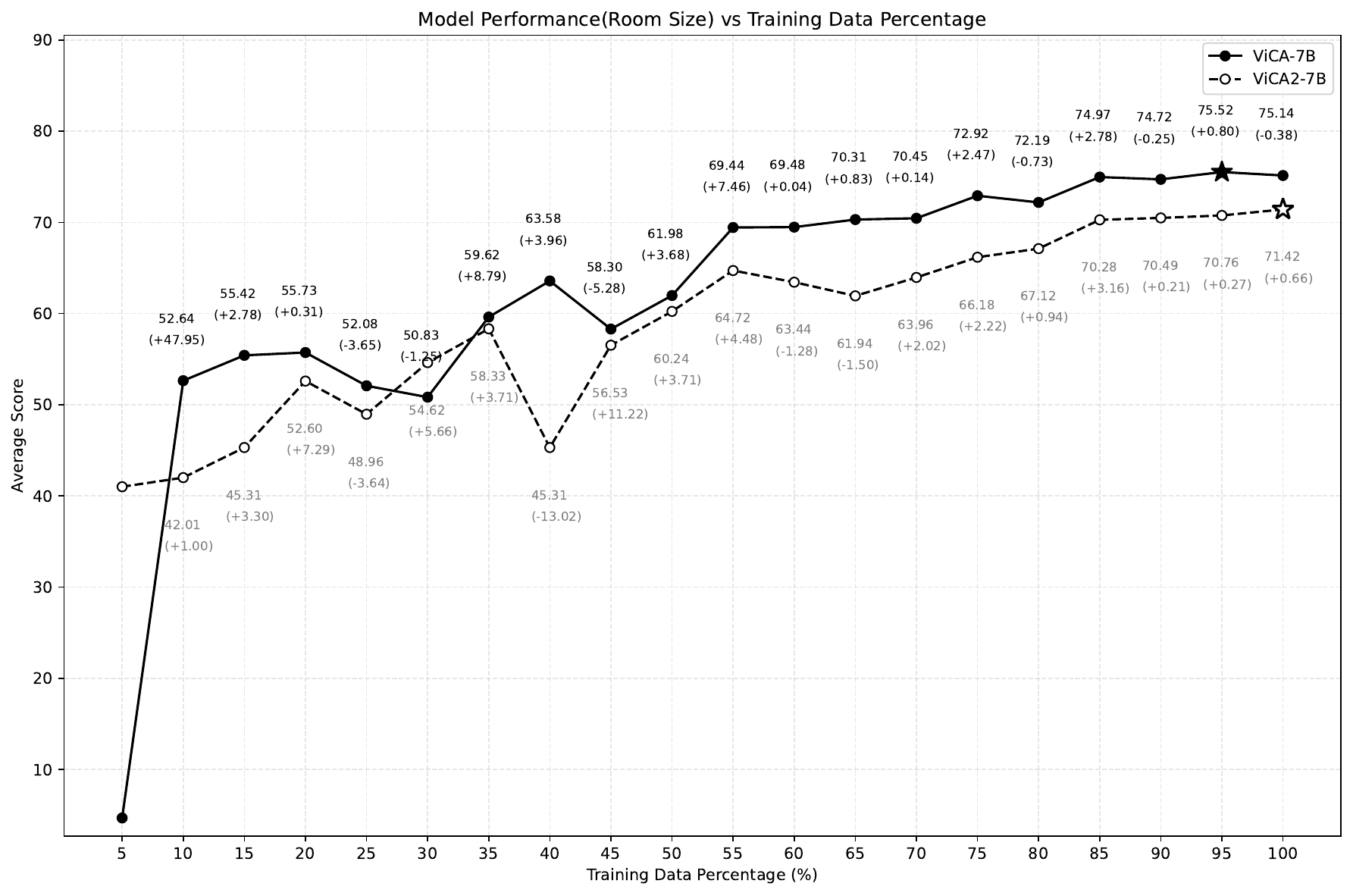}
  \caption{
  Performance Comparison of ViCA and ViCA2 with Increasing Training Data Size - \textbf{Room Size}.
  }
  \label{fig:scaling_comparison4}
\end{figure*}

\begin{figure*}[t]
  \centering
  \includegraphics[width=\textwidth]{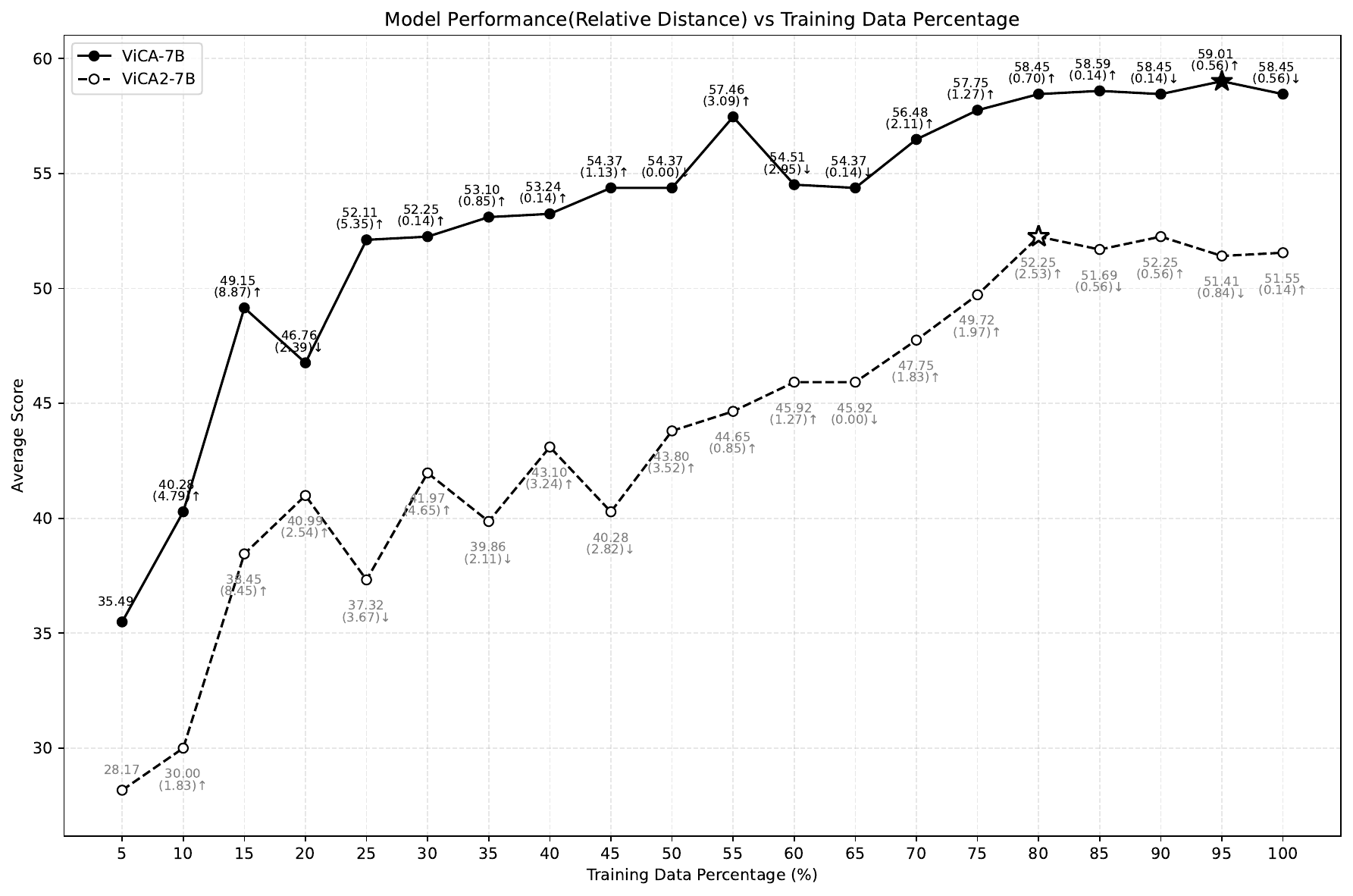}
  \caption{
  Performance Comparison of ViCA and ViCA2 with Increasing Training Data Size - \textbf{Relative Distance}.
  }
  \label{fig:scaling_comparison5}
\end{figure*}

\begin{figure*}[t]
  \centering
  \includegraphics[width=\textwidth]{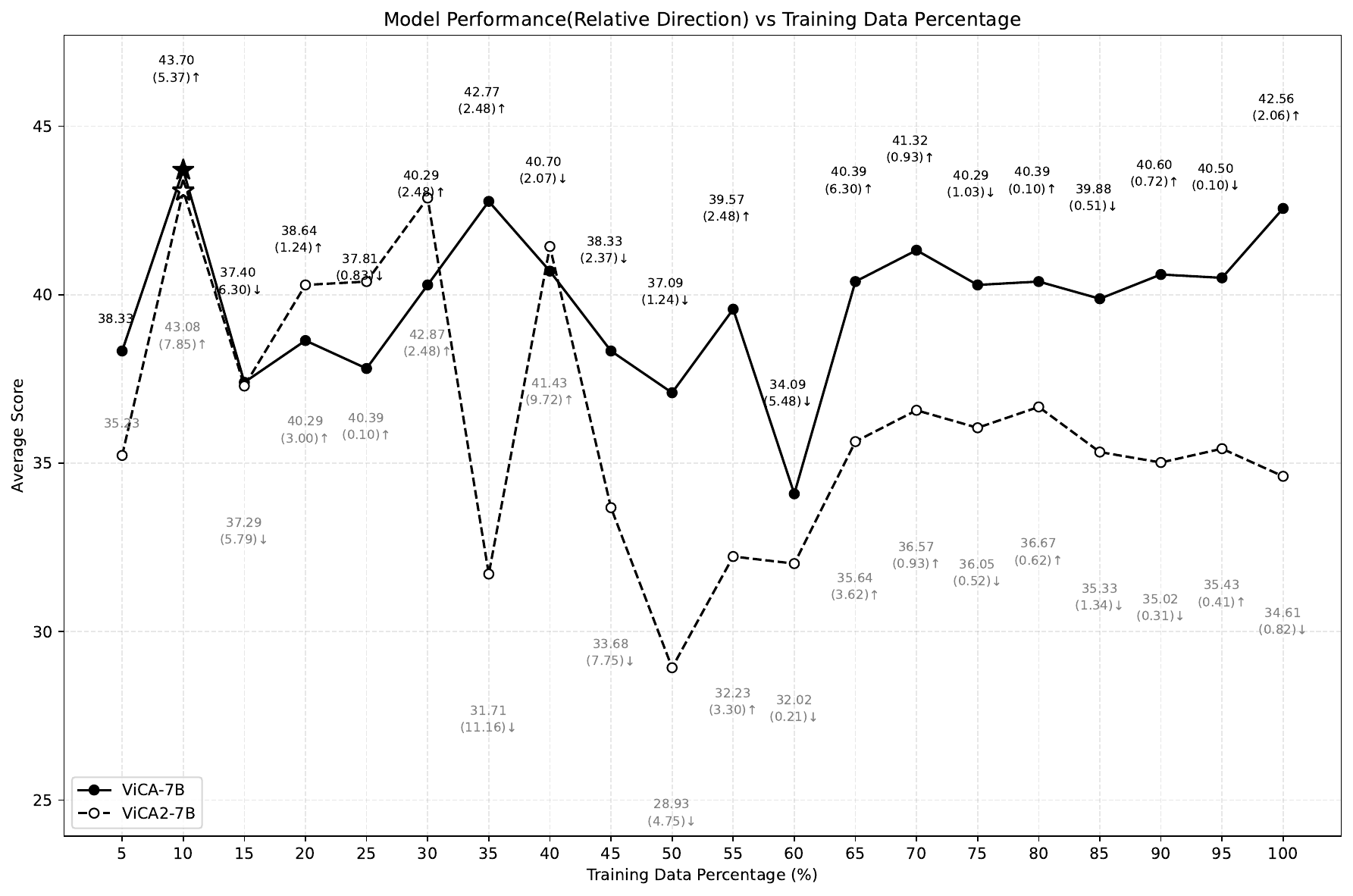}
  \caption{
  Performance Comparison of ViCA and ViCA2 with Increasing Training Data Size - \textbf{Relative Direction}.
  }
  \label{fig:scaling_comparison6}
\end{figure*}

\begin{figure*}[t]
  \centering
  \includegraphics[width=\textwidth]{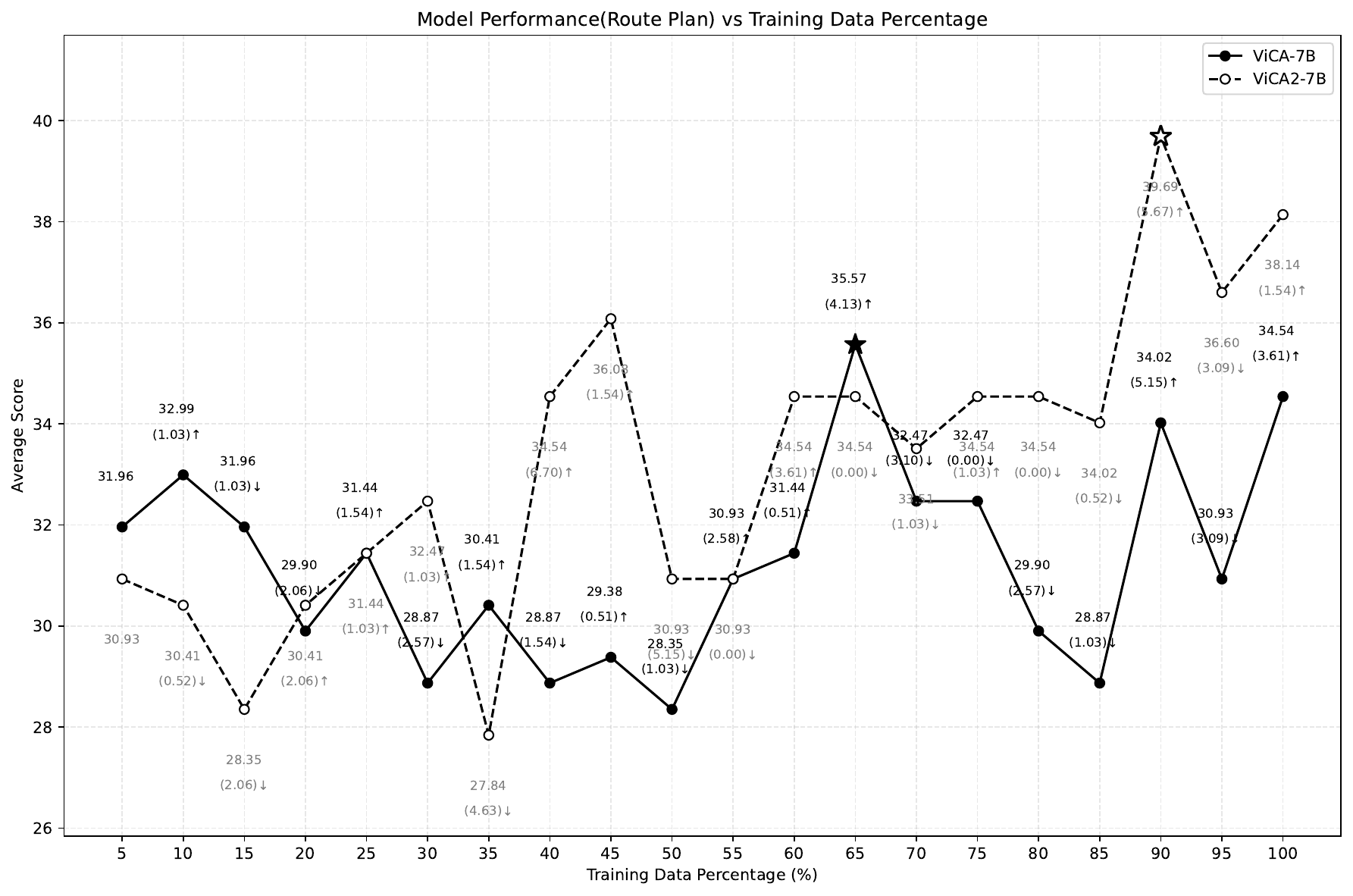}
  \caption{
  Performance Comparison of ViCA and ViCA2 with Increasing Training Data Size - \textbf{Route Plan}.
  }
  \label{fig:scaling_comparison7}
\end{figure*}

\begin{figure*}[t]
  \centering
  \includegraphics[width=\textwidth]{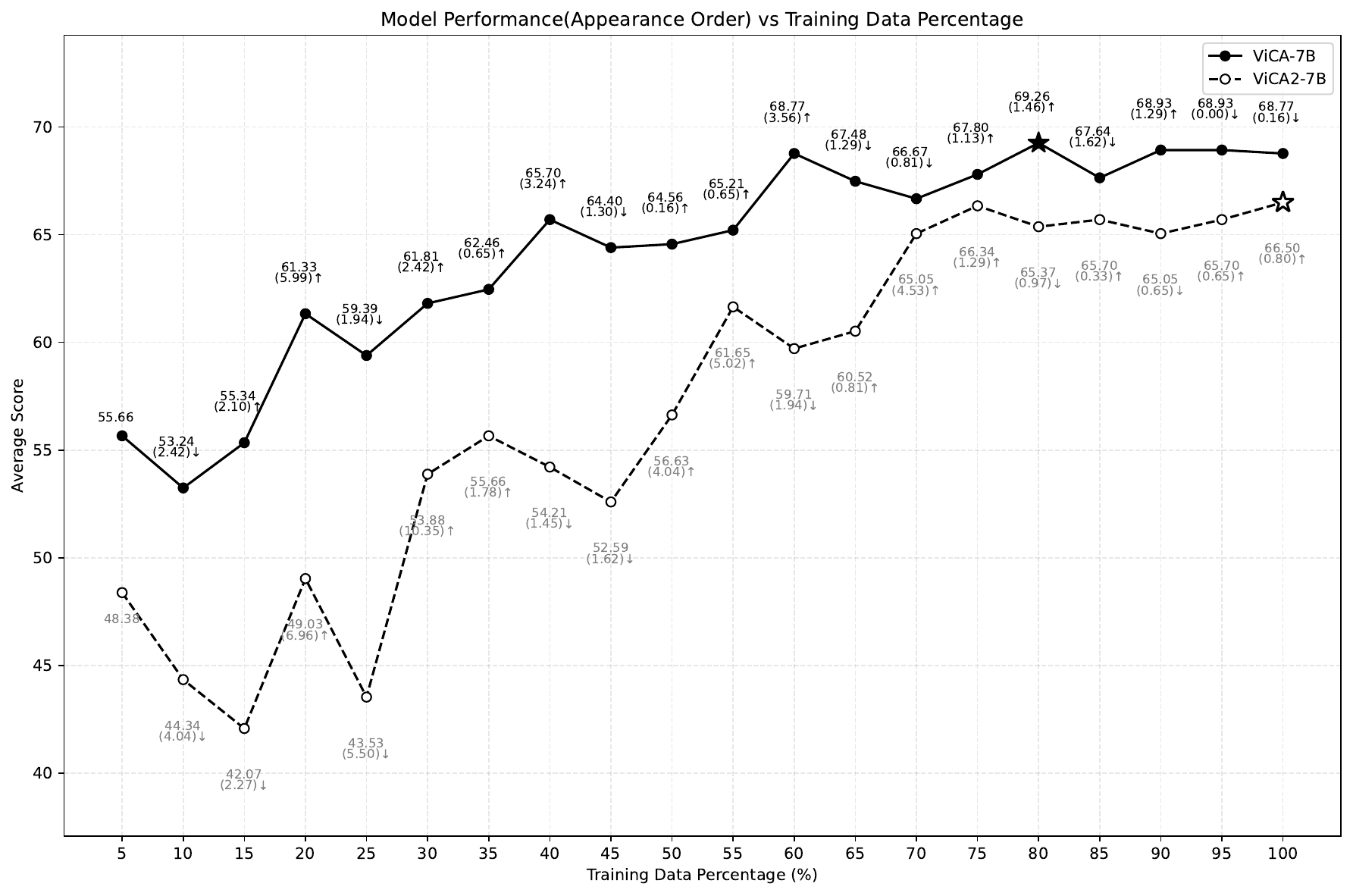}
  \caption{
  Performance Comparison of ViCA and ViCA2 with Increasing Training Data Size - \textbf{Appearance Order}.
  }
  \label{fig:scaling_comparison8}
\end{figure*}

To better understand how dataset size influences each model, we further analyzed the performance scaling curves of \textbf{ViCA-7B} and \textbf{ViCA2-7B} across all eight tasks in VSI-Bench. The results are presented in Figure~\ref{fig:scaling_comparison1}, \ref{fig:scaling_comparison2}, \ref{fig:scaling_comparison3}, \ref{fig:scaling_comparison4}, \ref{fig:scaling_comparison5}, \ref{fig:scaling_comparison6}, \ref{fig:scaling_comparison7}, \ref{fig:scaling_comparison8}.

We observe that for tasks well represented in the ViCA-322K dataset—such as \textit{Object Count}, \textit{Absolute Distance}, and \textit{Room Size}—both models exhibit clear performance gains as training data increases. In contrast, for tasks underrepresented or entirely absent from the dataset, namely \textit{Relative Direction} and \textit{Route Planning}, both models show unstable performance trends, with scores fluctuating without a clear upward trajectory. This suggests that the existing dataset lacks sufficient supervision for these tasks and that performance gains in these categories are unlikely without task-specific data augmentation.

These findings further support our earlier hypothesis: ViCA2-7B, with its greater architectural capacity, remains under-optimized under the current data regime and may benefit from scaling the dataset beyond the limits originally established for ViCA-7B.

\clearpage
\clearpage

\clearpage

\clearpage

\clearpage

\clearpage

\clearpage

\clearpage

\clearpage

\clearpage

\onecolumn
\section{Evaluation Prompts for Checkpoint Text Quality Across Training Stages}

The following are detailed descriptions of the same video, generated by four different Vision-Language Large Models (VLLMs). These models will be referred to as Model A, Model B, Model C, and Model D.

Your task is to critically evaluate and score the output from each of these VLLMs. The scoring scale is from 0 (minimum) to 10 (maximum).

Please ensure your evaluation addresses at least the following dimensions for each model's description:

1.  Richness of Detail: How comprehensive and specific are the details provided about the video's content?
2.  Accuracy: How accurately does the description reflect the presumed events, objects, and context within the video?
3.  Organization/Coherence : How logically structured, clear, and easy to follow is the description? Is there a coherent narrative or flow?
4.  Language Fluency: How natural, grammatically correct, and well-phrased is the language used?
5.  Information Redundancy or Repetition: Does the description contain unnecessary repetition of information or superfluous content?

After providing your detailed textual evaluation for each model, discussing its performance across these dimensions, please conclude by returning only the final scores in the precise JSON format exemplified below.

\begin{lstlisting}
json
{
  "A": <score_for_model_A>,
  "B": <score_for_model_B>,
  "C": <score_for_model_C>,
  "D": <score_for_model_D>
}
\end{lstlisting}

[A]\\

<Detailed descriptions of the video generated by model A>\\

[B]\\

<Detailed descriptions of the video generated by model B>\\

[C]\\

<Detailed descriptions of the video generated by model C>\\

[D]\\

<Detailed descriptions of the video generated by model D>\\

\twocolumn

\clearpage

\end{document}